%% file: main.tex
\documentclass{article} 
\usepackage{polpo_preprint,times}
\usepackage[T1]{fontenc}

\input{math_commands.tex}

\usepackage{amssymb}
\usepackage{amsthm}
\usepackage{booktabs}
\usepackage{graphicx}
\usepackage{xcolor}
\usepackage{float}
\usepackage{algorithm}
\usepackage{hyperref}
\hypersetup{pdftitle={Policy Optimization With Multiple Verifiers via Pareto Rankings},pdfauthor={Kai Yamashita, Shohei Taniguchi, Hiroki Furuta, Masahiro Suzuki, Yutaka Matsuo}}
\usepackage{xurl}
\usepackage{wrapfig}
\usepackage {colortbl}
\usepackage{placeins}
\definecolor{polpohighlight}{HTML}{ADD8E6}

\title{Policy Optimization with Multiple Verifiers \\ via Pareto Rankings}

\author{
\textbf{Kai Yamashita$^{1,*}$ \quad Shohei Taniguchi$^{1,2,*}$ \quad Hiroki Furuta} \\
\textbf{Masahiro Suzuki$^{1,2}$ \quad Yutaka Matsuo$^{1,2}$} \\[4pt]
\textnormal{$^{1}$The University of Tokyo \quad $^{2}$Third Intelligence, Inc.} \\
\textnormal{$^{*}$Equal contribution} \\
\textnormal{\texttt{kai.yamashita@weblab.t.u-tokyo.ac.jp}}
}

\DeclareMathOperator{\std}{std}
\DeclareMathOperator{\clip}{clip}
\DeclareMathOperator{\LE}{LE}
\DeclareMathOperator{\Max}{Max}
\DeclareMathOperator*{\lse}{logsumexp}
\newcommand{\RV}{\mathcal{R}_{V}}                 
\newcommand{\pPL}{p_{\theta}^{\mathrm{PL}}}       
\newcommand{\Arank}[1]{A_{#1}^{\mathrm{rank}}}    
\newcommand{\Agrpo}[1]{\widehat{A}_{#1}^{\mathrm{GRPO}}} 
\newcommand{\ind}[1]{\mathbf{1}\{#1\}}            
\newcommand{\eqn}[1]{Eq.~(\ref{#1})}             

\newtheorem{proposition}{Proposition}
\newtheorem{lemma}{Lemma}
\newtheorem{corollary}{Corollary}

\iclrfinalcopy 
\begin{document}

\maketitle

\begin{abstract}
Reinforcement learning with verifiable rewards (RLVR) has become an important approach to improving the reasoning capabilities of large language models. In tasks such as code generation and instruction following, multiple verifiers evaluate whether a response satisfies different requirements. Their outcomes provide detailed feedback about partial success, yet this feedback is commonly aggregated into a scalar reward for policy optimization. Rewarding only complete success discards distinctions among partial solutions, while counts and weighted sums impose comparisons between responses that satisfy different subsets of requirements. We introduce Partial-Order Likelihood Policy Optimization (POLPO), which learns from the comparisons established by verification without first constructing an aggregate reward. Pareto dominance identifies responses that satisfy additional requirements without losing any already satisfied, while leaving responses with conflicting outcomes incomparable. POLPO extends the listwise likelihood underlying Direct Preference Optimization by maximizing the total probability of all complete rankings consistent with these partial orders. This objective provides supervision from verified improvements even before complete solutions are available, without requiring a prescribed ordering of incomparable responses. Experiments on code generation and instruction following demonstrate improved learning efficiency and higher rates of complete requirement satisfaction compared with scalar-reward training. These results support partial-order likelihood as an effective approach to learning from multiple verifiers.
\end{abstract}

\section{Introduction}
\label{sec:intro}

Reinforcement learning with verifiable rewards (RLVR) has become an important approach to improving the reasoning capabilities of large language models (LLMs) \citep{lambert2024tulu3,deepseek2025r1}. By evaluating model-generated responses automatically, RLVR provides training supervision without requiring annotations for every intermediate reasoning step. In many applications, however, verification involves multiple requirements rather than a single correctness check. A generated program must pass tests covering different inputs and behaviors, while an instruction-following response may need to satisfy several content and formatting constraints \citep{hendrycks2021apps,pyatkin2025generalizing}. The resulting verification outcomes identify both whether a response succeeds completely and which requirements it satisfies along the way. How to use this information during training is therefore a central question for RLVR with multiple verifiers.

A common approach is to aggregate verification outcomes into a scalar reward and optimize the policy using a policy-gradient method, such as Group Relative Policy Optimization (GRPO) \citep{shao2024deepseekmath}. This aggregation determines which distinctions among responses are available to the learning algorithm. An all-pass reward directly expresses the goal of satisfying every requirement, but assigns the same reward to all partially successful responses. A pass-rate reward provides more graded feedback, but ranks responses by how many checks they pass, regardless of which checks they satisfy. Weighted sums allow different priorities, at the cost of specifying how gains on one requirement compensate for losses on another. Although these rewards can share complete success as their optimum, they induce different comparisons among the partial solutions encountered during training. The challenge is to learn from partial progress without having to prescribe trade-offs between requirements.

Partial orders offer a way to express this distinction. Under Pareto dominance, a response improves on another if it passes every check passed by the other and at least one additional check. Consider three programs $y_1,y_2,y_3$ with outcomes on three tests given by
\begin{equation}
  \mathbf c_1=(1,1,0),\qquad
  \mathbf c_2=(0,0,1),\qquad
  \mathbf c_3=(1,0,0),
  \label{eq:intro-checks}
\end{equation}
where $1$ denotes a passed test. Program $y_1$ improves on $y_3$, whereas $y_2$ passes a test that neither of the others passes. Verification therefore establishes
\begin{equation}
  y_1\succ_{\mathrm P}y_3,\qquad
  y_1\parallel_{\mathrm P}y_2,\qquad
  y_2\parallel_{\mathrm P}y_3,
  \label{eq:intro-pareto}
\end{equation}
where $\parallel_{\mathrm P}$ denotes incomparability. An all-pass reward misses the improvement from $y_3$ to $y_1$, while a pass-rate reward additionally favors $y_1$ over $y_2$. Treating both incomparabilities as scalar reward ties would instead force $y_1$ and $y_3$ to tie, erasing the verified improvement. A partial order preserves that improvement while leaving the trade-off involving $y_2$ unresolved.

\begin{figure}[t]
\centering
\includegraphics[width=\linewidth]{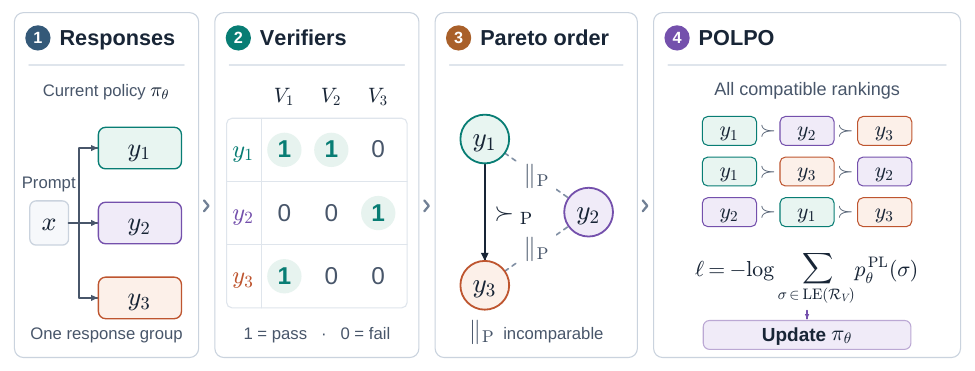}
\caption{Overview of POLPO. Responses sampled from the current rollout policy are checked against multiple requirements. In the Pareto-order panel, the solid arrow indicates verified dominance, while dashed connections indicate incomparability. For the verification outcomes in \eqn{eq:intro-checks}, only $y_1\succ y_3$ is established, so three complete rankings are compatible. POLPO minimizes the negative log of their total probability under the PL model parameterized by the policy scores in \eqn{eq:dpo-score}; the prompt and response group are omitted from the displayed objective for brevity.}
\label{fig:method}
\end{figure}

To turn these partial orders into a training objective, we introduce {\it Partial-Order Likelihood Policy Optimization (POLPO)}, illustrated in Figure~\ref{fig:method}. We use the implicit-reward parameterization of Direct Preference Optimization (DPO), originally proposed for aligning language models with human preferences \citep{rafailov2023dpo}. Policy-to-reference log probability ratios supply scores for a Plackett--Luce model, which assigns probabilities to rankings of the responses. POLPO maximizes the probability that all observed Pareto comparisons hold, obtained by summing the probabilities of the complete rankings consistent with them. For the partial order in \eqn{eq:intro-pareto}, every position for $y_2$ is allowed as long as $y_1$ precedes $y_3$. The objective thus captures the comparisons established by
verification while marginalizing over the ordering decisions
that verification leaves unspecified.
Our contribution is to bring the established partial-order PL
likelihood \citep{liu2019partial} into online policy learning
from multiple verifiers, using Pareto comparisons as supervision
and DPO-style scores to connect the likelihood to the response policy.

This formulation makes the partial order itself the supervision for policy optimization. Its likelihood models the specified comparisons jointly, and differentiating its logarithm yields a weighted sum of response log-probability gradients. Consequently, a group containing a Pareto improvement can provide a learning signal even when no response satisfies every requirement. Although POLPO uses scalar scores internally to parameterize ranking probabilities, it does not require a scalar reward that orders every pair of responses or represents incomparability through equal values. The distinction is between assigning scores within a probabilistic model and prescribing a complete ordering as the training target.

We evaluate whether this use of verification feedback improves learning in controlled program synthesis and practical code-generation tasks. The controlled task isolates the difficulty of learning from partial solutions that satisfy different subsets of requirements: POLPO reaches complete success with fewer sampled programs than training with either all-pass or pass-rate rewards. On MBPP and APPS, POLPO improves both complete program correctness and average test satisfaction under matched rollout budgets, compared with scalar-reward and comparison-based baselines. Training on AceCode further demonstrates transfer to external coding benchmarks. We also examine instruction following, where verifiers check heterogeneous requirements, to study how the approach generalizes beyond execution-based feedback. Together, these experiments investigate partial-order supervision as a way to use multiple verifiers directly in language-model training.

\section{Reinforcement Learning with Multiple Verifiers}
\label{sec:background}

In tasks with multiple requirements, a response can satisfy some requirements while violating others. We formalize this verification feedback and describe the scalar rewards commonly used to train policies from it.

\subsection{Task Setup}
\label{sec:rlvr}

Let $x\sim\rho$ denote a prompt and $\pi_\theta(y\mid x)$ a language-model policy. Each prompt has $K=K(x)\geq1$ binary verifiers, with $c_k(x,y)=1$ when response $y$ satisfies requirement $k$ and $0$ otherwise. Their outputs form a verification vector $\mathbf c(x,y)=(c_1(x,y),\ldots,c_K(x,y))\in\{0,1\}^K$. In code generation, each verifier can check a program on a test input; in instruction following, it can check a content or formatting constraint \citep{hendrycks2021apps,pyatkin2025generalizing}.

The task goal is to increase the probability of satisfying all requirements. The individual verifier outcomes also describe partial success: they identify which requirements are already satisfied when a response is not yet fully correct. For a group $Y=(y_1,\ldots,y_N)$ of responses to the same prompt, we write $\mathbf c_i=\mathbf c(x,y_i)$. The learning question is how to use these vectors to guide the policy toward complete solutions.

\subsection{Scalar-Reward Aggregation}
\label{sec:scalar-rewards}

A common approach maps the verification vector to a scalar reward $r(x,y)=g(\mathbf c(x,y))$ and optimizes its expectation:
\begin{equation}
  \mathcal J_{\mathrm{reward}}(\theta)
  =\E_{x\sim\rho,\,y\sim\pi_\theta(\cdot\mid x)}[r(x,y)].
  \label{eq:rlvr-objective}
\end{equation}
For example, an all-pass reward is one only when every check passes, whereas a pass-rate reward averages the individual outcomes:
\begin{equation}
  r_{\mathrm{all}}(x,y)=\prod_{k=1}^{K}c_k(x,y),
  \qquad
  r_{\mathrm{rate}}(x,y)=\frac{1}{K}\sum_{k=1}^{K}c_k(x,y).
  \label{eq:reward-baselines}
\end{equation}
An all-pass reward directly expresses complete success, but gives every partially successful response the same reward. A pass-rate reward distinguishes partial solutions by the number of checks passed, while assigning equal rewards to responses that pass different subsets of the same size. A weighted sum, $r_{\mathbf w}(x,y)=\sum_{k=1}^{K}w_k c_k(x,y)$ with $w_k>0$, permits different priorities but specifies how gains on one check compensate for losses on another. These choices can favor the same fully successful responses while assigning different rewards to the partial solutions encountered during training.

The aggregation determines the verification information supplied to the policy optimizer. GRPO is one representative policy-gradient method that uses this scalar feedback \citep{shao2024deepseekmath}. It samples several responses per prompt and normalizes their rewards within the group to obtain advantages for a policy update. If every response receives the same reward, as with all-pass rewards when no response succeeds completely, the reward-derived advantages vanish. See Appendix~\ref{app:grpo-gradient} for the advantage definition and policy surrogate.

These limitations motivate using the improvements established by verification directly as comparison targets, while leaving tradeoffs between conflicting outcomes unspecified.

\section{Partial-Order Likelihood Policy Optimization}
\label{sec:method}

\subsection{Pareto Comparisons from Multiple Verifiers}
\label{sec:partial-orders}

To use verification feedback without prescribing trade-offs between requirements, we compare responses through Pareto dominance. A response dominates another if it passes every check passed by the other and at least one additional check. Writing $\mathbf c_i=(c_{i1},\ldots,c_{iK})$, we define
\begin{equation}
y_i\succ_{\mathrm P}y_j
\quad\Longleftrightarrow\quad
\bigl[\forall k,\ c_{ik}\geq c_{jk}\bigr]
\ \land\
\bigl[\exists k,\ c_{ik}>c_{jk}\bigr].
\label{eq:pareto}
\end{equation}
When each response passes a check that the other fails, neither dominates the other. This leaves the tradeoff between their satisfied requirements unspecified. Responses with identical verification vectors are also left unordered. For the verification outcomes in \eqn{eq:intro-checks}, verification establishes only $y_1\succ_{\mathrm P}y_3$; both pairs involving $y_2$ remain incomparable, as shown in \eqn{eq:intro-pareto}.

The resulting relation on response indices, $\RV=\{(i,j):y_i\succ_{\mathrm P}y_j\}$, is a strict partial order: it is irreflexive and transitive. With a single binary verifier, every correct response precedes every incorrect response, with no comparisons within either group. More generally, we write $\RV=\mathcal A(\mathbf c_1,\ldots,\mathbf c_N)$ for a task-specific comparison rule, requiring that its output be a strict partial order; Pareto dominance is the rule used for multi-verifier feedback here. An unordered pair supplies no comparison target and does not require equal model scores. The remaining question is how to turn this partial-order supervision into a policy-training objective.

\subsection{Learning from the Partial-Order Likelihood}
\label{sec:likelihood}

To turn these Pareto comparisons into a policy-training objective, we build on the formulation of Direct Preference Optimization (DPO), which was originally proposed for aligning language models with human preferences \citep{rafailov2023dpo}. DPO uses an implicit reward derived from the policy to model comparison probabilities. We use this parameterization to model the probability that all comparisons in $\RV$ hold jointly. Given a fixed reference policy $\pi_{\mathrm{ref}}$, we define the response score as
\begin{equation}
  s_\theta(x,y)
  =\beta\log\frac{\pi_\theta(y\mid x)}{\pi_{\mathrm{ref}}(y\mid x)}.
  \label{eq:dpo-score}
\end{equation}
Here $\beta>0$ sets the score scale, and both log probabilities sum over response tokens (see Appendix~\ref{app:sequence-logprob}). The implicit-reward interpretation follows from KL-regularized reward maximization: holding $\theta$ fixed, $s_\theta$ is a reward for which $\pi_\theta$ is the optimal policy,
\begin{equation}
\pi_\theta(\cdot\mid x)
=\operatorname*{arg\,max}_{q}
\Bigl\{
\E_{y\sim q}[s_\theta(x,y)]
-\beta D_{\mathrm{KL}}\!\left(q\,\Vert\,\pi_{\mathrm{ref}}(\cdot\mid x)\right)
\Bigr\},
\label{eq:implicit-reward-optimum}
\end{equation}
where $q$ ranges over distributions on complete responses. Adding a prompt-dependent constant $C(x)$ to the reward preserves this optimizer and cancels from ranking probabilities. See Appendix~\ref{app:implicit-reward} for a derivation of this correspondence. POLPO learns the implicit reward through the comparison likelihood below: the verification vectors determine which comparisons are observed, while the policy scores determine their probabilities.

We use the Plackett--Luce (PL) model \citep{luce1959individual,plackett1975analysis} to define a distribution over rankings from these scores. Write $s_i=s_\theta(x,y_i)$ and let $\sigma=(\sigma_1,\ldots,\sigma_N)$ list responses from best to worst. The probability of this ranking is
\begin{equation}
\pPL(\sigma\mid x,Y)
=\prod_{k=1}^{N-1}
\frac{\exp(s_{\sigma_k})}
{\sum_{j=k}^{N}\exp(s_{\sigma_j})}.
\label{eq:pl}
\end{equation}
At each position, a response is selected with probability proportional to its exponentiated score among the remaining candidates; the final choice has probability one. Higher implicit rewards therefore make a response more likely to appear earlier in the ranking.

To see how this ranking model can learn from partial comparisons, return to the example in \eqn{eq:intro-pareto}. Verification establishes only $y_1\succ y_3$, leaving the position of $y_2$ unspecified. There are exactly three complete rankings consistent with this feedback:
\begin{equation*}
y_1\succ y_2\succ y_3,\qquad
y_1\succ y_3\succ y_2,\qquad
y_2\succ y_1\succ y_3.
\end{equation*}
Each ranking satisfies the verified comparison, so learning should increase their total probability without selecting a particular position for $y_2$. Since the three rankings are mutually exclusive, their probabilities add up to the probability that $y_1$ precedes $y_3$:
\begin{equation}
\begin{aligned}
p_\theta(\{(1,3)\}\mid x,Y)
&=\pPL((1,2,3)\mid x,Y)+\pPL((1,3,2)\mid x,Y)+\pPL((2,1,3)\mid x,Y)\\
&=\frac{e^{s_1}}{e^{s_1}+e^{s_3}}.
\end{aligned}
\label{eq:pareto-three}
\end{equation}
The score of $y_2$ cancels because it participates in no verified comparison. The learning signal therefore depends only on the comparison between $y_1$ and $y_3$. Summing over the allowed positions of $y_2$ preserves the unspecified part of the order, without requiring equal scores for incomparable responses.

For a general partial order $R_V$, we adopt the established
Plackett--Luce likelihood obtained by summing over its linear
extensions \citep{liu2019partial}.
These are the complete rankings that satisfy every comparison in
$R_V$, and we denote their set by $\mathrm{LE}(R_V)$.
A ranking belongs to this set exactly when $y_i$ precedes $y_j$
for every $(i,j)\in R_V$.
POLPO uses this likelihood to train the response policy:
verification determines $R_V$, the policy-to-reference scores in
\eqn{eq:dpo-score} parameterize the ranking probabilities, and training
minimizes the negative log-likelihood:
\begin{equation}
\ell_{\mathrm{POLPO}}(\theta;x,Y,\RV)
=-\log p_\theta(\RV\mid x,Y)
=-\log\sum_{\sigma\in\LE(\RV)}\pPL(\sigma\mid x,Y).
\label{eq:polpo}
\end{equation}
This sum marginalizes over the ordering decisions left unspecified by verification. The scalar scores parameterize the ranking probabilities; the partial order supplies the comparison targets. A complete ranking leaves a single PL term, while an empty comparison relation gives likelihood one and zero loss. The objective recovers DPO for a single comparison (see Appendix~\ref{app:dpo}), but generally differs from a sum of pairwise DPO losses when multiple comparisons are observed (see Appendix~\ref{app:marginalization}).

We optimize \eqn{eq:polpo} with gradient-based updates to the policy parameters, differentiating through the scores in \eqn{eq:dpo-score}. The likelihood is evaluated exactly using a dynamic program over subsets of responses, avoiding explicit enumeration of complete rankings. In our coding experiments, computing this loss instead of GRPO's objective adds less than 0.1\% to the training time (Appendix~\ref{app:cost}). In instruction-following settings, the corresponding initial-policy loss overhead is 0.03--0.12\% of the reported training-update time. See Appendices~\ref{app:derivations} and~\ref{app:implementation} for the gradient derivations and likelihood computation, and see Algorithm~\ref{alg:polpo} for the training procedure.

\section{Experiments}
\label{sec:experiments}

We study whether comparisons from multiple verifiers help a policy learn to satisfy every requirement.
A controlled program-synthesis task isolates learning from partial solutions, coding tests this idea with a language model, and multi-instruction following extends it to heterogeneous requirements.
The main comparison is between POLPO and GRPO with an all-pass reward (GRPO--All) or the fraction of passed checks (GRPO--Rate).
These scalar rewards are defined in \eqn{eq:reward-baselines}.
Complete success is the primary outcome; average check satisfaction is a complementary measure of partial progress.
See Appendix~\ref{app:experimental-details} for the task-specific training and evaluation settings.

\subsection{Controlled Program Synthesis}
\label{sec:exp-toy}

We first isolate learning from multiple verifier outcomes using the controlled FizzBuzz synthesis task in Figure~\ref{fig:toy-task}.
For divisors $(a,b)$, the target returns \texttt{Fizz}, \texttt{Buzz}, or \texttt{FizzBuzz} according to divisibility and otherwise returns the input itself; standard FizzBuzz uses $(a,b)=(3,5)$.
A categorical policy constructs a fixed three-branch program by choosing three divisor--output pairs $(d_j,v_j)$, with $d_j\in\{2,\ldots,M\}$ and $v_j\in\{\texttt{Fizz},\texttt{Buzz},\texttt{FizzBuzz}\}$.
Each input $n\in\{1,\ldots,60\}$ defines a binary verifier, giving every sampled program a 60-dimensional verification vector.
Because the policy is small, we can compute its exact probability of generating a program that passes all verifiers.

\begin{figure}[H]
\centering
\begingroup
\newcommand{\modelslot}[1]{\colorbox{blue!12}{\strut\texttt{#1}}}
\setlength{\fboxsep}{5pt}
\fbox{%
\begin{minipage}{0.94\linewidth}
\small
\textbf{Model output: six categorical choices}
\[
  (d_1,v_1),\ (d_2,v_2),\ (d_3,v_3),
  \qquad
  d_j\in\{2,\ldots,M\},\quad
  v_j\in\{\texttt{Fizz},\texttt{Buzz},\texttt{FizzBuzz}\}.
\]
\vspace{-0.8em}
\begin{minipage}[t]{0.48\linewidth}
\vspace{0pt}
\textbf{Synthesized program}
\par\vspace{0.4em}
\scriptsize
\begin{tabular}{@{}l@{}}
\texttt{program(n):} \\
\texttt{\ \ if n \% }\modelslot{$d_1$}\texttt{ == 0: return }\modelslot{$v_1$} \\
\texttt{\ \ if n \% }\modelslot{$d_2$}\texttt{ == 0: return }\modelslot{$v_2$} \\
\texttt{\ \ if n \% }\modelslot{$d_3$}\texttt{ == 0: return }\modelslot{$v_3$} \\
\texttt{\ \ return str(n)}
\end{tabular}
\end{minipage}\hfill
\begin{minipage}[t]{0.48\linewidth}
\vspace{0pt}
\textbf{Verifiers}
\par\vspace{0.4em}
\scriptsize
\begingroup
\renewcommand{\arraystretch}{1.65}
\begin{tabular}{@{}l@{}}
\texttt{for n = 1, ..., 60:} \\
\texttt{\ \ expected = target}$_{a,b}$\texttt{(n)} \\
\texttt{\ \ if program(n) == expected:} \\
\texttt{\ \ \ \ }$c_n\leftarrow 1$\texttt{\ \ \# pass} \\
\texttt{\ \ else: }$c_n\leftarrow 0$\texttt{\ \ \# fail}
\end{tabular}
\endgroup
\par\vspace{0.6em}
\noindent$\mathbf c(\texttt{program})=(c_1,\ldots,c_{60})\in\{0,1\}^{60}.$
\end{minipage}
\par\vspace{0.5em}
\end{minipage}}
\endgroup
\caption{Controlled program-synthesis task.
The trainable policy fills only the highlighted branch slots; the program structure, fallback, target outputs, and 60 input-specific verifiers are fixed.
For $(a,b)=(3,5)$, the two correct assignments are $(15,\texttt{FizzBuzz})$ followed by $(3,\texttt{Fizz})$ and $(5,\texttt{Buzz})$ in either order.}
\label{fig:toy-task}
\end{figure}

GRPO--All receives no signal from incomplete programs, whereas GRPO--Rate orders them only by the number of passed inputs.
POLPO instead preserves componentwise dominance between their verification vectors.
We also include two variants with the same likelihood objective: POLPO--Rate ranks programs by mean verifier satisfaction, leaving equal-rate programs unordered, and POLPO--All places completely correct programs above all others, leaving the rest unordered.
These comparisons isolate the choice of supervision while keeping the policy, reference, and optimizer fixed.
We test whether this distinction reduces the number of sampled programs needed to reach a $0.9$ probability of complete success, varying both the divisor search space $M$ and the target divisors $(a,b)$.
See Appendix~\ref{app:toy-details} for the complete protocol.

\begin{figure}[H]
\centering
\includegraphics[width=0.8\linewidth]{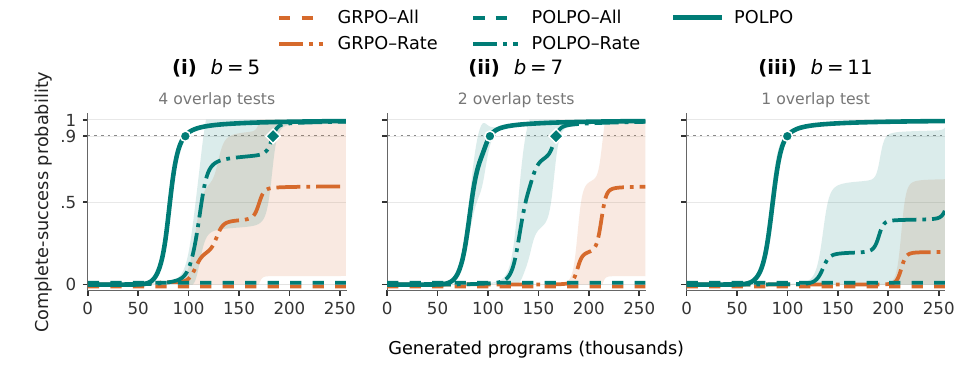}
\caption{Complete-success probability for target pairs $(3,5)$, $(3,7)$, and $(3,11)$ at $M=60$, averaged over five training seeds.
All settings retain the same 60 input verifiers and initial policy.
Shading shows one sample standard deviation across seeds, clipped to $[0,1]$.
The horizontal dotted line marks $0.9$, and markers indicate the first recorded crossings of the mean curves.}
\label{fig:toy-learning}
\end{figure}

\paragraph{Results.}
Across all three target pairs in Figure~\ref{fig:toy-learning}, POLPO reaches $0.9$ complete-success probability in all five seeds, using approximately 96,000--100,000 programs on average.
Within the full budget of 512,000 programs, GRPO--All and POLPO--All never reach the threshold, while GRPO--Rate reaches it in three, three, and one of the five seeds for $(3,5)$, $(3,7)$, and $(3,11)$, respectively.
POLPO--Rate requires more samples than POLPO for the first two targets and fails to reach the threshold in two seeds for $(3,11)$.
POLPO--All obtains no comparison until a completely correct program is sampled; in smaller search spaces, where such programs are sampled, it closely follows GRPO--All and remains slower than POLPO (Appendix~\ref{app:toy-efficiency}).
Thus, the choice of comparison supervision matters even with the same likelihood objective.
For $(3,11)$ at 153,600 programs, GRPO--Rate has 98.07\% mean verifier satisfaction but only a 0.03\% probability of passing the overlap input $n=33$, compared with 99.28\% for POLPO, averaged across seeds.
Thus, a high scalar pass rate can conceal a persistently unsatisfied part of the specification; see Appendix~\ref{app:toy-efficiency} for threshold counts and diagnostic curves.

\subsection{Coding Tasks}
\label{sec:exp-coding}

Code generation provides a direct setting for learning from multiple verifiers, with each test checking a particular program behavior.
We study MBPP, APPS Introductory, and transfer from AceCode to external benchmarks.
All methods use the same tests and eight fresh responses per prompt, with one optimizer update per rollout batch.
We evaluate final checkpoints using greedy decoding. MBPP results report the mean and standard deviation over three training seeds; APPS Introductory and AceCode results use one training seed.
For MBPP, \emph{All-pass} measures the percentage of problems whose response passes every test (pass@1), while \emph{Avg. pass} averages the fraction of tests passed within each problem, then equally across problems.
See Appendix~\ref{app:coding-details} for the full protocols.

\paragraph{Baselines.}
GRPO--All and GRPO--Rate provide scalar-reward baselines.
On MBPP and APPS, we additionally include REINFORCE++--All and REINFORCE++--Rate \citep{hu2025reinforcepp}, which use the same rewards with batch-wide normalization to test scalar-reward learning beyond GRPO's group-relative update.
GDPO \citep{gdpo} instead normalizes each test's outcomes separately before aggregation, providing a baseline that explicitly uses individual verifier signals.
DPO--Pareto applies the DPO objective \citep{rafailov2023dpo} to all strict Pareto dominance pairs, while GraphDPO \citep{liu2026graphdpo} learns local rankings over each response and those it dominates.
These two baselines use the same Pareto relation as POLPO, allowing us to compare pairwise and local graph objectives with the likelihood of the entire partial order.
For MBPP, we also include POLPO--Rate, which uses the same partial-order likelihood objective and hyperparameters as POLPO but constructs comparisons from mean test-pass rates, leaving equal-rate responses unordered.
This ablation isolates the effect of Pareto-based supervision while retaining the likelihood objective.
POLPO--All from Section~\ref{sec:exp-toy} also retains the likelihood objective but compares only responses that pass every test with the others, whereas GRPO--Pareto uses the Pareto relation without the likelihood objective, taking as its advantages in GRPO's update the coefficients of POLPO's gradient at equal scores (Appendix~\ref{app:coding-details}).

\paragraph{MBPP and APPS Introductory.}
MBPP contains short Python programming problems with three tests per problem \citep{austin2021mbpp}.
We train Qwen3-0.6B and Qwen3-4B with thinking enabled, using the official training and test splits.
POLPO achieves the highest All-pass and Avg. pass at both model sizes (Table~\ref{tab:mbpp-coding}), improving complete program correctness as well as average test satisfaction.
Compared with POLPO--Rate, POLPO improves All-pass from 42.00\% to 44.73\% and Avg.\ pass from 48.53\% to 52.42\% for Qwen3-0.6B.
For Qwen3-4B, the corresponding improvements are from 67.40\% to 67.93\% and from 71.93\% to 75.49\%.
These results suggest that Pareto-based comparisons provide more informative supervision than pass-rate rankings when the likelihood objective and hyperparameters are held fixed.
POLPO--All and GRPO--Pareto remain below POLPO and do not exceed GRPO--All or GRPO--Rate on either metric for either model, so POLPO's improvement appears to require both the Pareto relation and the partial-order likelihood.


\begin{table}[t]
\centering
\small
\caption{MBPP test results (\%) after five training epochs.
Both models use thinking during training and evaluation, with one greedy response per problem.
Trained-model results are reported as mean $\pm$ standard deviation over three training seeds; the initial model is shown without a seed standard deviation.
All-pass measures success on every test (pass@1); Avg. pass averages the fraction of tests passed equally across problems.}
\label{tab:mbpp-coding}
\begin{tabular}{lrrrr}
\toprule
& \multicolumn{2}{c}{Qwen3-0.6B} & \multicolumn{2}{c}{Qwen3-4B} \\
\cmidrule(lr){2-3}\cmidrule(lr){4-5}
Method & All-pass & Avg. pass & All-pass & Avg. pass \\
\midrule
Initial model & 30.00 & 34.40 & 62.40 & 66.40 \\
GRPO--All & $41.47 \pm 0.95$ & $46.96 \pm 1.23$ & $66.53 \pm 0.31$ & $71.49 \pm 0.89$ \\
GRPO--Rate & $42.47 \pm 0.64$ & $48.60 \pm 0.50$ & $66.60 \pm 0.53$ & $71.18 \pm 0.69$ \\
GRPO--Pareto & $40.53 \pm 0.42$ & $46.11 \pm 0.60$ & $66.20 \pm 0.35$ & $71.00 \pm 0.27$ \\
REINFORCE++--All & $40.33 \pm 1.67$ & $46.89 \pm 0.20$ & $64.80 \pm 1.74$ & $69.56 \pm 2.05$ \\
REINFORCE++--Rate & $38.47 \pm 1.14$ & $45.02 \pm 1.58$ & $64.47 \pm 1.17$ & $68.87 \pm 1.03$ \\
GDPO & $36.47 \pm 1.40$ & $41.76 \pm 1.99$ & $66.60 \pm 0.53$ & $70.98 \pm 0.15$	 \\
GraphDPO & $41.27 \pm 1.42$ & $47.40 \pm 1.03$ & $65.87 \pm 0.99$ & $69.89 \pm 1.61$ \\
DPO--Pareto & $39.73 \pm 2.30$ & $45.84 \pm 2.12$ & $66.27 \pm 0.76$ & $71.07 \pm 1.01$ \\
POLPO--All & $39.93 \pm 1.70$ & $45.24 \pm 2.28$ & $66.40 \pm 0.69$ & $70.73 \pm 0.24$ \\
POLPO--Rate & $42.00 \pm 1.91$ & $48.53 \pm 1.84$ & $67.40 \pm 2.11$ & $71.93 \pm 2.03$ \\
\rowcolor{polpohighlight}
POLPO & $\mathbf{44.73} \pm 0.61$ & $\mathbf{52.42} \pm 1.00$ & $\mathbf{67.93} \pm 0.83$ & $\mathbf{75.49} \pm 0.56$ \\
\bottomrule
\end{tabular}
\end{table}

APPS provides competition-style programming problems with a variable number of tests \citep{hendrycks2021apps}.
We train and evaluate Qwen2.5-Coder-1.5B-Instruct on its Introductory tier.
POLPO leads both metrics across the compared methods (Table~\ref{tab:apps-coding}), extending the improvement to a setting where the number and coverage of verifiers vary across problems.


\paragraph{AceCode and external evaluation.}
To assess transfer beyond the training problems, we train the same Qwen2.5-Coder model on AceCode's hard subset \citep{zeng2025acecoder} and evaluate on HumanEval and the MBPP subset distributed with EvalPlus \citep{liu2023evalplus}.
We report pass@1 on both the original and augmented test suites; this MBPP subset differs from the official test split used above.
POLPO achieves the highest pass@1 on all four evaluations, tied with GRPO--All on original HumanEval (Table~\ref{tab:acecode-coding}).


\begin{table}[t]
\centering
\small
\setlength{\tabcolsep}{4pt}
\begin{minipage}[t]{0.45\textwidth}
\vspace{0pt}
\centering
\caption{APPS Introductory test results (\%) after two training epochs.}
\label{tab:apps-coding}
\begin{tabular}{lrr}
\toprule
Method & All-pass & Avg. pass \\
\midrule
Initial model & 20.50 & 35.77 \\
GRPO--All & 23.60 & 39.05 \\
GRPO--Rate & 23.30 & 38.95 \\
REINFORCE++--All & 24.30 & 38.75 \\
REINFORCE++--Rate & 23.20 & 38.36 \\
GDPO & 23.40 & 39.13 \\
GraphDPO & 21.80 & 37.38 \\
DPO--Pareto & 21.70 & 37.57 \\
\rowcolor{polpohighlight}
POLPO & \textbf{25.90} & \textbf{42.67} \\
\bottomrule
\end{tabular}
\end{minipage}\hfill
\begin{minipage}[t]{0.53\textwidth}
\vspace{0pt}
\centering
\caption{External pass@1 (\%) after one epoch of AceCode training. + denotes augmented tests.}
\label{tab:acecode-coding}
\begin{tabular}{lrrrr}
\toprule
& \multicolumn{2}{c}{HumanEval (164)} & \multicolumn{2}{c}{MBPP (378)} \\
\cmidrule(lr){2-3}\cmidrule(lr){4-5}
Method & Original & + & Original & + \\
\midrule
Initial model & 70.12 & 65.24 & 68.52 & 59.52 \\
GRPO--All & \textbf{78.05} & 70.73 & 73.02 & 62.96 \\
GRPO--Rate & 76.83 & 70.73 & 73.02 & 62.43 \\
GraphDPO & 75.61 & 68.90 & 71.16 & 62.17 \\
DPO--Pareto & 75.61 & 68.90 & 71.16 & 61.90 \\
\rowcolor{polpohighlight}
POLPO & \textbf{78.05} & \textbf{72.56} & \textbf{74.87} & \textbf{64.29} \\
\bottomrule
\end{tabular}
\end{minipage}
\end{table}

\subsection{Multi-Instruction Following}
\label{sec:exp-multi}

We evaluate learning from multiple verifiable constraints on IFM5, the IF-RLVR data released by \citet{pyatkin2025generalizing}, MulDimIF \citep{ye2025muldimif}, and MathIF \citep{fu2025scaling}. IFM5 and MulDimIF use separate models trained on their respective training splits and evaluated on the corresponding test splits. MathIF additionally evaluates mathematical correctness alongside instruction adherence, using models trained on a separate constraint-augmented MATH dataset.
For IFM5 and MulDimIF, the initial model is Qwen3-4B-Instruct-2507. On both datasets, we compare the initial model, GRPO--All, GRPO--Rate, GDPO \citep{gdpo}, GraphDPO, DPO--Pareto, and POLPO. The primary metric is the percentage of evaluated prompts whose responses satisfy every constraint. Appendix~\ref{app:if-details} provides the data splits, evaluation protocols, and training configurations.

\begin{table*}[t]
\centering
\small
\setlength{\tabcolsep}{2.5pt}
\caption{Complete constraint satisfaction (\%) on IFM5. $K$ counts individual constraints.
Trained-model results are reported as mean $\pm$ SD over three seeds; the initial model is shown without a seed standard deviation.
Bold marks column maxima at the reported precision.}
\label{tab:ifm5-overall}

\begin{tabular}{@{}lrrrrrr@{}}
\toprule
Method & $K=1$ & $K=2$ & $K=3$ & $K=4$ & $K=5$ & Overall \\
\midrule
Initial model & 56.1 & 29.9 & 18.5 & 7.4 & 3.7 & 27.1 \\
GRPO--All & $90.7\pm0.4$ & $65.9\pm0.5$ & $45.8\pm1.1$ & $28.6\pm1.1$ & $19.8\pm2.6$ & $56.4\pm0.8$ \\
GRPO--Rate & $90.1\pm0.6$ & $66.0\pm2.7$ & $48.7\pm2.3$ & $28.8\pm2.4$ & $20.5\pm1.1$ & $57.1\pm1.7$ \\
GDPO & $89.1\pm0.6$ & $65.8\pm1.0$ & $48.3\pm0.7$ & $29.4\pm0.3$ & $20.7\pm1.3$ & $56.8\pm0.3$ \\
GraphDPO & $91.2\pm0.5$ & $69.0\pm2.5$ & $52.0\pm3.4$ & $31.0\pm3.0$ & $24.4\pm2.7$ & $59.6\pm2.2$ \\
DPO--Pareto & $\mathbf{92.4}\pm0.2$ & $69.3\pm0.4$ & $53.4\pm1.7$ & $\mathbf{34.6}\pm0.9$ & $25.2\pm3.0$ & $61.0\pm0.6$ \\
\rowcolor{polpohighlight}
POLPO & $92.1\pm0.6$ & $\mathbf{69.7}\pm0.2$ & $\mathbf{54.9}\pm1.4$ & $33.8\pm0.0$ & $\mathbf{26.2}\pm3.5$ & $\mathbf{61.3}\pm0.2$ \\
\bottomrule
\end{tabular}

\end{table*}

\begin{table*}[t]
\centering
\small
\setlength{\tabcolsep}{2.5pt}
\caption{Complete constraint satisfaction (\%) on MulDimIF. L1--L4 are difficulty levels.
Trained-model results are reported as mean $\pm$ SD over three seeds; the initial model is shown without a seed standard deviation.
Bold marks column maxima at the reported precision.}
\label{tab:muldimif-overall}
\begin{tabular}{@{}lrrrrr@{}}
\toprule
Method & L1 & L2 & L3 & L4 & Overall \\
\midrule
Initial model & 85.3 & 63.7 & 48.0 & 32.3 & 57.3 \\
GRPO--All & $95.3\pm0.3$ & $89.7\pm0.3$ & $85.3\pm0.3$ & $83.0\pm1.5$ & $88.3\pm0.2$ \\
GRPO--Rate & $95.2\pm0.7$ & $89.2\pm0.8$ & $86.6\pm1.3$ & $85.6\pm2.4$ & $89.1\pm1.0$ \\
GDPO & $95.3\pm1.2$ & $89.4\pm0.8$ & $85.4\pm0.4$ & $84.1\pm1.3$ & $88.6\pm0.5$ \\
GraphDPO & $96.4\pm0.4$ & $89.8\pm1.2$ & $86.4\pm1.7$ & $\mathbf{86.2}\pm0.5$ & $89.7\pm0.8$ \\
DPO--Pareto & $96.0\pm1.2$ & $89.9\pm0.5$ & $86.7\pm2.0$ & $83.4\pm1.1$ & $89.0\pm0.4$ \\
\rowcolor{polpohighlight}
POLPO & $\mathbf{96.6}\pm1.2$ & $\mathbf{90.6}\pm0.7$ & $\mathbf{87.4}\pm2.0$ & $85.7\pm2.1$ & $\mathbf{90.1}\pm0.7$ \\
\bottomrule
\end{tabular}
\end{table*}

\paragraph{IFM5 and MulDimIF.}
IFM5 groups prompts by their number of constraints, $K\in\{1,\ldots,5\}$, whereas MulDimIF defines four difficulty levels based on both the number and variety of constraints \citep{ye2025muldimif}. Table~\ref{tab:ifm5-overall} shows that POLPO has the highest overall mean on IFM5 (61.3\%), closely followed by DPO--Pareto (61.0\%). POLPO leads at $K=2,3,5$, while DPO--Pareto leads at $K=1,4$. On MulDimIF (Table~\ref{tab:muldimif-overall}), POLPO has the highest overall mean (90.1\%) and leads at L1--L3; GraphDPO leads at L4. These are comparisons of means across three training seeds.

\begin{table}[t]
\centering
\small
\setlength{\tabcolsep}{3pt}
\caption{MathIF results (\%) on 420 examples by thinking mode, with a 16,384-token evaluation response limit. Bold indicates the best result among the methods shown within each mode. Dashes indicate results not reported for that mode.}
\label{tab:mathif-overall}
\begin{tabular}{@{}lrrrrrrrr@{}}
\toprule
& \multicolumn{4}{c}{Thinking disabled} & \multicolumn{4}{c}{Thinking enabled} \\
\cmidrule(lr){2-5}\cmidrule(lr){6-9}
Method & HAcc & SAcc & Acc & Joint & HAcc & SAcc & Acc & Joint \\
\midrule
Initial model & 31.2 & 55.7 & 13.8 & 5.7 & 25.2 & 40.6 & 25.0 & 6.2 \\
GRPO--All & 55.7 & 74.3 & \textbf{31.4} & 20.2 & 41.9 & 46.5 & 27.1 & 22.6 \\
GRPO--Rate & 76.0 & 86.0 & 27.1 & 19.3 & 70.5 & 79.3 & 38.8 & 30.7 \\
GDPO & -- & -- & -- & -- & 58.6 & 68.4 & 30.5 & 23.3 \\
GraphDPO & -- & -- & -- & -- & 56.4 & 63.4 & 30.0 & 24.8 \\
DPO--Pareto & -- & -- & -- & -- & 72.6 & 82.7 & \textbf{40.2} & 31.9 \\
\rowcolor{polpohighlight}
POLPO & \textbf{81.4} & \textbf{89.7} & 26.9 & \textbf{21.4} & \textbf{78.1} & \textbf{86.1} & \textbf{40.2} & \textbf{33.8} \\
\bottomrule
\end{tabular}
\end{table}
\paragraph{MathIF.}
We further evaluate whether responses can satisfy instruction constraints while solving mathematical problems on MathIF \citep{fu2025scaling}. We evaluate Qwen3-0.6B on 420 MathIF examples with thinking disabled and enabled. HAcc measures complete constraint satisfaction, SAcc the mean fraction of constraints satisfied, and Acc mathematical correctness. Joint, the primary metric, requires both mathematical correctness and satisfaction of every instruction constraint. Among the methods shown in Table~\ref{tab:mathif-overall}, POLPO achieves the highest Joint score and instruction adherence in both modes. With thinking disabled, GRPO--All has the highest mathematical accuracy (31.4\%), while POLPO has the highest Joint score (21.4\%). With thinking enabled, POLPO and DPO--Pareto tie for the highest mathematical accuracy (40.2\%); POLPO has the highest Joint score (33.8\%), followed by DPO--Pareto (31.9\%). Appendices~\ref{app:mathif-details} and~\ref{app:mathif-results} provide details and breakdowns. Appendix~\ref{app:if-rollout-analysis} examines learning-signal availability and Pareto-order structure in instruction-following rollouts.

\section{Related Work}
\label{sec:related}

\paragraph{Multi-verifier RLVR and reward design.}
Multiple verification outcomes provide training supervision in RLVR tasks, such as instruction following and code generation \citep{pyatkin2025generalizing,zeng2025acecoder}. A central design choice is how to turn partial success into a policy-training signal. Pass-rate rewards provide denser feedback than all-pass rewards, but do not consistently improve complete-solution accuracy in code generation \citep{li2026passrate}. GDPO normalizes reward components separately before aggregation \citep{gdpo}, while VeRPO constructs execution-based rewards that account for partial test success, test difficulty, and test density \citep{wang2026verpo}. In multi-instruction following, Scarcity-Aware Rewards counter exploration biased toward easier instructions by assigning rewards according to how rarely each instruction is satisfied \citep{zhang2026mitigating}. HiR takes a complementary approach: it reuses partially successful responses by rewriting their prompts to retain the instructions they satisfy \citep{zhang2025replay}. POLPO uses verification outcomes to identify supported comparisons between responses and learns their joint likelihood.

Vector-valued feedback also supports learning across competing objectives. Vector Policy Optimization trains diverse sets of responses that specialize in different reward trade-offs, improving their usefulness for test-time search \citep{vpo}. ParetoPO targets the trade-off between task accuracy and tool-use efficiency through hypervolume-guided reward modulation and Pareto-ranking-based advantages \citep{li2026paretopo}. In our setting, the goal is to produce a response satisfying all verified requirements. Pareto dominance identifies improvements among partially successful responses without specifying how gains on one requirement should compensate for losses on another.

\paragraph{Preference optimization with structured comparisons.}
DPO parameterizes comparison probabilities through policy-to-reference log-probability ratios \citep{rafailov2023dpo}. IRPO and DPO-VP apply DPO-based objectives to correct--incorrect response pairs, establishing a connection between verification and preference learning \citep{pang2024irpo,tu2025iterativedpo}. PRO and LiPO extend preference optimization to ranked response lists \citep{song2024pro,liu2025lipo}. Listwise Policy Optimization also learns from response groups in RLVR, using scalar rewards to construct target distributions for policy projection \citep{qu2026lpo}. GraphDPO represents response comparisons as a directed acyclic graph and aggregates local choice losses, each contrasting a response with the responses it dominates \citep{liu2026graphdpo}. It therefore shares POLPO's use of structured comparisons without requiring a complete ordering. The distinction lies in the objective: GraphDPO aggregates choice probabilities over overlapping neighborhoods, whereas POLPO optimizes the probability that all observed comparisons hold jointly in a single Plackett--Luce ranking. For general partial orders, these local losses do not constitute a factorization of the joint ranking-event likelihood.

Plackett--Luce models have also been used to learn from supervision that leaves some ranking decisions unspecified. \citet{ma2021partitioned} study the marginalized likelihood of ordered partitions, where the order between groups is observed but the order within each group is unknown. In recommendation, Mult-DPO considers comparisons between sets of positive and negative items and derives a tractable multinomial surrogate that upper-bounds the corresponding marginalized PL loss \citep{zhu2026multdpo}. POLPO builds on this probabilistic treatment of incomplete rankings. It uses verifier-induced Pareto comparisons to define the likelihood of a general partial order and optimizes that likelihood through policy-derived scores. See Appendix~\ref{app:extended-related} for additional related work.

\section{Conclusion}
\label{sec:conclusion}

We studied how multiple verification outcomes can guide policy learning toward satisfying all requirements. POLPO represents verified improvements as Pareto comparisons and maximizes the probability that these comparisons hold jointly, marginalizing over unspecified ordering decisions. This formulation enables learning from partial success without prescribing trade-offs between requirements. In controlled program synthesis, POLPO reaches a high probability of complete success with fewer sampled programs than training with all-pass or pass-rate rewards. Coding and multi-instruction-following experiments further show improvements in complete requirement satisfaction, with coding gains extending to external benchmarks. These findings support partial-order supervision as an effective approach to learning from multiple verifiers.

\paragraph{Limitation.}
POLPO's learning signal depends on the comparisons available within each response group. When no response strictly Pareto-dominates another, the group provides no comparison signal, even if some requirements remain unsatisfied. This can limit learning when sampled responses satisfy different, conflicting subsets of requirements. An important direction for future work is to develop sampling strategies that expose verified improvements and provide more informative comparison supervision.

\subsection*{AI use statement}

In this work, we used generative AI tools (Claude Code and Codex) to clean and reformat datasets, support qualitative and thematic data analysis, assist in the writing of proofs, formulate mathematical claims, provide critical ingredients for proving mathematical claims, implement methods, and assist with translation. We have not used generative AI tools to generate synthetic datasets, propose or refine hypotheses, design or provide feedback on research methodology or experiments, or interpret results, and the development of theoretical models or conceptual frameworks is not applicable to this work. Additionally, we used generative AI tools to support literature reviews, create scientific figures, assist with drafting parts of the research paper, summarize and analyse existing literature, brainstorm ideas, source and search for information, and edit the research paper to improve readability. We have reviewed all AI-assisted work. The author personally reviewed the AI-assisted mathematical claims and proofs, method implementations, data processing and qualitative analyses, literature summaries, translations, figures, and manuscript text. We take responsibility for the final content of this work, including text, claims or artifacts produced with the aid of generative AI.

\subsection*{Ethics statement}

This work studies a training objective for language models
using existing benchmark datasets and programmatically
constructed tasks. Automatic verifiers assess only
specified requirements and may contain errors; satisfying
these checks does not guarantee that generated responses
are safe, fair, or appropriate for deployment. Improvements
in code generation and instruction following may also
facilitate harmful applications. Models trained with this
objective therefore require safety evaluation and safeguards
appropriate to their intended use.

\subsection*{Reproducibility statement}

Section~\ref{sec:background} defines the verification outcomes and scalar rewards.
Section~\ref{sec:method} defines the comparison rules, policy scores, and training objective.
See Appendix~\ref{app:derivations} for the derivations and proofs for the likelihood and its gradients.
See Appendix~\ref{app:implementation} for the training algorithm, exact likelihood computation, and the conditions under which response groups are included in training.
See Appendix~\ref{app:experimental-details} for the datasets, preprocessing, training configurations, and evaluation procedures.

\bibliography{main}
\bibliographystyle{iclr2027_conference}
\clearpage

\appendix

\section{Policy Optimization Preliminaries}
\label{app:preliminaries}

This appendix specifies the sequence-log-probability convention and GRPO surrogate, derives the correspondence between implicit rewards and KL-regularized optimal policies, and shows how the partial-order likelihood recovers DPO for a single comparison.

\subsection{Sequence Log Probabilities}
\label{app:sequence-logprob}

For a response of length $T_i$, we use the sequence log probability
\begin{equation}
\log\pi_\theta(y_i\mid x)
=\sum_{t=1}^{T_i}\log\pi_\theta(y_{i,t}\mid x,y_{i,<t}),
\label{eq:app-seq-logprob}
\end{equation}
where $y_{i,<t}$ denotes preceding response tokens. All policy scores in this paper use this sum over tokens.

\subsection{Group Relative Policy Optimization}
\label{app:grpo-gradient}

GRPO samples a response group from a rollout policy $\pi_{\mathrm{old}}$, a snapshot held fixed during the ensuing updates \citep{shao2024deepseekmath}. Given scalar rewards $\mathbf{r}=(r_1,\ldots,r_N)$, it computes
\begin{equation}
\bar r=\frac{1}{N}\sum_{j=1}^{N}r_j,
\qquad
\Agrpo{i}
=\frac{r_i-\bar r}{\std(\mathbf{r})+\delta},
\label{eq:app-grpo-advantage}
\end{equation}
where $\std(\mathbf{r})$ is the group standard deviation and $\delta>0$ is a stabilizer. Each token in $y_i$ receives the same advantage: responses above the group mean have positive advantages, and those below it have negative advantages. GRPO uses these weights in a clipped policy surrogate, with an additional KL penalty in its original formulation. The surrogate and its gradient are given below.

The policy update uses the token probability ratio
\begin{equation}
q_{i,t}(\theta)
=\frac{\pi_\theta(y_{i,t}\mid x,y_{i,<t})}
{\pi_{\mathrm{old}}(y_{i,t}\mid x,y_{i,<t})}.
\label{eq:grpo-ratio}
\end{equation}
For one response group, the clipped policy surrogate is
\begin{equation}
\begin{split}
\mathcal{J}_{\mathrm{GRPO}}^{\mathrm{clip}}(\theta;x,Y)
=\frac{1}{N}\sum_{i=1}^{N}\frac{1}{T_i}
\sum_{t=1}^{T_i}
\min\Bigl\{
&\,q_{i,t}(\theta)\,\Agrpo{i},\\
&\,\clip\bigl(q_{i,t}(\theta),1-\varepsilon,1+\varepsilon\bigr)\,\Agrpo{i}
\Bigr\},
\end{split}
\label{eq:grpo-surrogate}
\end{equation}
where $\varepsilon>0$ controls clipping, as in proximal policy optimization \citep{schulman2017ppo}. GRPO maximizes this surrogate averaged over prompts and rollout groups, with an additional KL penalty to a reference policy in the original formulation. The comparison with POLPO concerns the clipped policy term alone (see Appendix~\ref{app:po-gradient}).

\begin{proposition}
\label{prop:grpo-gradient}
At $\theta=\theta_{\mathrm{old}}$, the gradient of the clipped surrogate \eqn{eq:grpo-surrogate} is
\begin{equation}
\left.\nabla_\theta
\mathcal{J}_{\mathrm{GRPO}}^{\mathrm{clip}}(\theta;x,Y)
\right|_{\theta=\theta_{\mathrm{old}}}
=\frac{1}{N}\sum_{i=1}^{N}
\frac{\Agrpo{i}}{T_i}
\left.\nabla_\theta\log\pi_\theta(y_i\mid x)
\right|_{\theta=\theta_{\mathrm{old}}}.
\label{eq:grpo-gradient}
\end{equation}
\end{proposition}

\begin{proof}
The advantages $\Agrpo{i}$ depend only on the rewards and are constants. At $\theta=\theta_{\mathrm{old}}$ every ratio equals $q_{i,t}(\theta_{\mathrm{old}})=1$, which lies strictly inside the clipping interval $[1-\varepsilon,1+\varepsilon]$. In a neighborhood of this point the clipping function is the identity, so both arguments of the minimum in \eqn{eq:grpo-surrogate} coincide with $q_{i,t}(\theta)\Agrpo{i}$, and the surrogate is differentiable with
\begin{equation*}
\nabla_\theta\bigl[q_{i,t}(\theta)\Agrpo{i}\bigr]
=\Agrpo{i}\,q_{i,t}(\theta)\,\nabla_\theta\log\pi_\theta(y_{i,t}\mid x,y_{i,<t}).
\end{equation*}
Evaluating at $\theta_{\mathrm{old}}$ sets $q_{i,t}=1$. Summing over tokens and using the sequence log probability \eqn{eq:app-seq-logprob} gives
$\frac{1}{N}\sum_{i}\bigl(\Agrpo{i}/T_i\bigr)\nabla_\theta\log\pi_\theta(y_i\mid x)$ at $\theta_{\mathrm{old}}$, which is \eqn{eq:grpo-gradient}.
\end{proof}

\subsection{Implicit Rewards and KL-Regularized Optimal Policies}
\label{app:implicit-reward}

For a fixed prompt $x$ and reward function $r(x,y)$, consider the KL-regularized objective
\begin{equation}
\mathcal J_r(q\mid x)
=\E_{y\sim q}[r(x,y)]
-\beta D_{\mathrm{KL}}\!\left(q\,\Vert\,\pi_{\mathrm{ref}}(\cdot\mid x)\right),
\label{eq:app-kl-reward-objective}
\end{equation}
with $\beta>0$. The distributions are over complete responses, rather than just the sampled group $Y$. Assume that the reference and policies under consideration have common positive support, the partition function below is finite, and the displayed expectations and divergences are finite. Define
\begin{equation}
\pi_r^*(y\mid x)
=\frac{\pi_{\mathrm{ref}}(y\mid x)\exp(r(x,y)/\beta)}{Z_r(x)},
\qquad
Z_r(x)=\sum_y\pi_{\mathrm{ref}}(y\mid x)\exp(r(x,y)/\beta).
\label{eq:app-gibbs-policy}
\end{equation}
Substituting this expression into \eqn{eq:app-kl-reward-objective} gives
\begin{equation}
\mathcal J_r(q\mid x)
=\beta\log Z_r(x)
-\beta D_{\mathrm{KL}}\!\left(q\,\Vert\,\pi_r^*(\cdot\mid x)\right).
\label{eq:app-kl-reward-identity}
\end{equation}
The reward is fixed while optimizing over $q$, so $Z_r(x)$ is constant. Nonnegativity of KL divergence shows that the unique maximizing response distribution is $q=\pi_r^*(\cdot\mid x)$. Rearranging \eqn{eq:app-gibbs-policy} yields the inverse reward parameterization used in DPO \citep{rafailov2023dpo}:
\begin{equation}
r(x,y)
=\beta\log\frac{\pi_r^*(y\mid x)}{\pi_{\mathrm{ref}}(y\mid x)}
+\beta\log Z_r(x).
\label{eq:app-inverse-reward}
\end{equation}
The final term depends only on the prompt and cancels from comparison probabilities.

Conversely, fix $\theta$ and set $r_\theta^{\mathrm{imp}}(x,y)=s_\theta(x,y)+C(x)$, using the score in \eqn{eq:dpo-score} and any finite prompt-dependent $C(x)$. Then $Z_{r_\theta^{\mathrm{imp}}}(x)=\exp(C(x)/\beta)$ and $\pi_{r_\theta^{\mathrm{imp}}}^*=\pi_\theta$, since $\sum_y\pi_\theta(y\mid x)=1$. Equivalently,
\begin{equation}
\mathcal J_{r_\theta^{\mathrm{imp}}}(q\mid x)
=C(x)-\beta D_{\mathrm{KL}}\!\left(q\,\Vert\,\pi_\theta(\cdot\mid x)\right),
\label{eq:app-implicit-reward-identity}
\end{equation}
which proves \eqn{eq:implicit-reward-optimum} by taking $C(x)=0$. This establishes the policy--reward correspondence for each fixed $\theta$. During POLPO training, $\theta$ changes through comparison-likelihood optimization, so the associated implicit reward changes with it; the identity does not assert maximization of a fixed scalar verifier reward. Rescaling the reward and KL coefficient by the same positive factor preserves the optimal policy. Taking $C(x)=0$ and dividing both by $\beta$ gives an equivalent unit-KL interpretation with reward $\log(\pi_\theta/\pi_{\mathrm{ref}})$; the PL likelihood continues to use the score $s_\theta=\beta\log(\pi_\theta/\pi_{\mathrm{ref}})$.

\subsection{Connection to Direct Preference Optimization}
\label{app:dpo}

For a group containing two responses $Y=(y_w,y_l)$ and the comparison $y_w\succ y_l$, the partial-order event contains only the ranking that places $y_w$ first. Writing $s_w=s_\theta(x,y_w)$ and $s_l=s_\theta(x,y_l)$ using \eqn{eq:dpo-score}, its PL probability is the Bradley--Terry comparison probability \citep{bradley1952rank}:
\begin{equation}
  p_\theta^{\mathrm{pair}}(y_w\succ y_l\mid x)
  =\frac{e^{s_w}}{e^{s_w}+e^{s_l}}
  =\frac{1}{1+\exp[-(s_w-s_l)]}.
  \label{eq:bt}
\end{equation}
Taking the negative logarithm gives the DPO loss \citep{rafailov2023dpo},
\begin{equation}
  \ell_{\mathrm{DPO}}(\theta;x,y_w,y_l)
  =-\log p_\theta^{\mathrm{pair}}(y_w\succ y_l\mid x)
  =\log\bigl(1+\exp[-(s_w-s_l)]\bigr),
  \label{eq:dpo-loss}
\end{equation}
where $s_w-s_l=\beta\bigl[\log(\pi_\theta(y_w\mid x)/\pi_{\mathrm{ref}}(y_w\mid x))-\log(\pi_\theta(y_l\mid x)/\pi_{\mathrm{ref}}(y_l\mid x))\bigr]$. A prompt-dependent additive constant in the implicit reward cancels from this difference and, more generally, from every PL choice probability. Thus the score in \eqn{eq:dpo-score} can be used directly in the likelihood.

The same pairwise likelihood results when a larger response group has only one specified comparison, by Lemma~\ref{lem:pairwise}. If every position is specified, the partial-order likelihood contains a single PL term and recovers the listwise extension of DPO \citep[Appendix~A.3]{rafailov2023dpo}. Verification-derived comparison targets, such as correct--incorrect response pairs used by DPO-based reasoning methods \citep{pang2024irpo,tu2025iterativedpo}, fit within this likelihood formulation.

\section{Derivations and Proofs}
\label{app:derivations}

This appendix collects the derivations behind Section~\ref{sec:method}. Throughout, $Y=(y_1,\ldots,y_N)$ is a fixed response group, $s_i=s_\theta(x,y_i)$ are the policy scores in \eqn{eq:dpo-score}, $w_i=e^{s_i}$, and $\RV$ is a strict partial order on $\{1,\ldots,N\}$ with linear extensions $\LE(\RV)$. For a complete ranking $\sigma$, $R_k(\sigma)=\{\sigma_k,\ldots,\sigma_N\}$ denotes the candidates remaining at position $k$, and
\begin{equation}
p_{i}^{(k)}(\sigma)=\frac{w_i}{\sum_{j\in R_k(\sigma)}w_j}
\qquad\text{for }i\in R_k(\sigma)
\label{eq:app-choice-prob}
\end{equation}
is the probability of choosing $y_i$ at position $k$ given the remaining candidates. Set $p_i^{(k)}(\sigma)=0$ for $i\notin R_k(\sigma)$.

\subsection{Gradient of the Plackett--Luce Log-Likelihood}
\label{app:pl-gradient}

\begin{proposition}[Complete-ranking coefficients]
\label{prop:pl-gradient}
For every complete ranking $\sigma$ and every $i\in\{1,\ldots,N\}$,
\begin{equation}
a_i(\sigma)=\frac{\partial}{\partial s_i}\log\pPL(\sigma\mid x,Y)
=\sum_{k=1}^{N-1}\Bigl[\ind{i=\sigma_k}-\ind{i\in R_k(\sigma)}\,p_i^{(k)}(\sigma)\Bigr],
\label{eq:pl-coefficient}
\end{equation}
where $\ind{\cdot}$ is the indicator function. Moreover, $\sum_{i=1}^{N}a_i(\sigma)=0$.
\end{proposition}

\begin{proof}
By \eqn{eq:pl},
$\log\pPL(\sigma\mid x,Y)=\sum_{k=1}^{N-1}\bigl[s_{\sigma_k}-\log\sum_{j\in R_k(\sigma)}e^{s_j}\bigr]$.
The derivative of the $k$-th term with respect to $s_i$ is $\ind{i=\sigma_k}$ from the first part and $-\ind{i\in R_k(\sigma)}\,e^{s_i}/\sum_{j\in R_k(\sigma)}e^{s_j}$ from the log-partition function, which gives the stated expression. Summing over $i$, position $k$ contributes $1-\sum_{i\in R_k(\sigma)}p_i^{(k)}(\sigma)=0$, because the choice probabilities at each position sum to one over the remaining candidates.
\end{proof}

Since $i=\sigma_k$ implies $i\in R_m(\sigma)$ exactly for $m\leq k$, the coefficient of the response at position $k$ can also be written as $a_{\sigma_k}(\sigma)=1-\sum_{m=1}^{k}p_{\sigma_k}^{(m)}(\sigma)$ for $k\leq N-1$ and $a_{\sigma_N}(\sigma)=-\sum_{m=1}^{N-1}p_{\sigma_N}^{(m)}(\sigma)$ for the last response. For $k<N$, the coefficient is positive exactly when $\sum_{m=1}^{k}p_{\sigma_k}^{(m)}(\sigma)<1$, zero when this sum equals one, and negative when it exceeds one. For positive weights and $N\geq 2$, the final response has a strictly negative coefficient. The sum here combines choice probabilities from different positions; it is not a probability that the response has already been selected.

\subsection{Gradient of the Partial-Order Likelihood}
\label{app:po-gradient}

\begin{proposition}[Partial-order coefficients]
\label{prop:po-gradient}
Let $p_\theta(\RV\mid x,Y)=\sum_{\sigma\in\LE(\RV)}\pPL(\sigma\mid x,Y)$ as in \eqn{eq:polpo}. Then
\begin{equation}
\Arank{i}=\frac{\partial}{\partial s_i}\log p_\theta(\RV\mid x,Y)
=\sum_{\sigma\in\LE(\RV)}\omega_\theta(\sigma)\,a_i(\sigma),
\qquad
\omega_\theta(\sigma)=\frac{\pPL(\sigma\mid x,Y)}{p_\theta(\RV\mid x,Y)},
\label{eq:rank-coefficient-mixture}
\end{equation}
where the weights $\omega_\theta(\sigma)$ are the conditional probabilities of the compatible rankings given that the comparisons hold and sum to one. Consequently $\sum_{i=1}^{N}\Arank{i}=0$, and
\begin{equation}
\nabla_\theta\log p_\theta(\RV\mid x,Y)
=\beta\sum_{i=1}^{N}\Arank{i}\,\nabla_\theta\log\pi_\theta(y_i\mid x).
\label{eq:polpo-gradient}
\end{equation}
\end{proposition}

\begin{proof}
The set $\LE(\RV)$ does not depend on $\theta$, so
$\partial_{s_i}p_\theta(\RV\mid x,Y)=\sum_{\sigma\in\LE(\RV)}\partial_{s_i}\pPL(\sigma\mid x,Y)
=\sum_{\sigma\in\LE(\RV)}\pPL(\sigma\mid x,Y)\,a_i(\sigma)$,
using $\partial_{s_i}\pPL=\pPL\,\partial_{s_i}\log\pPL$. Dividing by $p_\theta(\RV\mid x,Y)$ gives the mixture form, and the weights sum to one by definition. The zero-sum property follows from Proposition~\ref{prop:pl-gradient}: $\sum_i\Arank{i}=\sum_\sigma\omega_\theta(\sigma)\sum_i a_i(\sigma)=0$. For the policy gradient, the reference policy is fixed, so $\nabla_\theta s_i=\beta\,\nabla_\theta\log\pi_\theta(y_i\mid x)$ by \eqn{eq:dpo-score}, and the chain rule gives
$\nabla_\theta\log p_\theta(\RV\mid x,Y)=\sum_{i}\bigl(\partial_{s_i}\log p_\theta(\RV\mid x,Y)\bigr)\nabla_\theta s_i$.
\end{proof}

The likelihood thus supplies a coefficient for each response gradient, depending on both the verified comparisons and the current policy scores. These coefficients are not standardized advantages and need not have unit variance; their scale comes from the likelihood. Their zero-sum property also follows from the invariance of ranking probabilities to a common shift of all scores.

\paragraph{Connection to GRPO updates.}
Both the POLPO gradient in \eqn{eq:polpo-gradient} and the GRPO surrogate gradient at $\theta=\theta_{\mathrm{old}}$ in \eqn{eq:grpo-gradient} weight response gradients using relative information from the group. GRPO derives advantages from scalar rewards; POLPO derives its coefficients from the joint comparison likelihood. This is a connection between update structures, not an equivalence of objectives. GRPO averages token contributions and uses clipping, whereas POLPO uses sequence log probabilities and directly optimizes the likelihood, with no additional explicit KL term in \eqn{eq:polpo}. See Appendix~\ref{app:grpo-gradient} for the full GRPO surrogate and the conditions for \eqn{eq:grpo-gradient}.

\subsection{Marginalization over Unspecified Positions}
\label{app:marginalization}

Two facts about the Plackett--Luce model explain the marginalization in \eqn{eq:pareto-three} and the additional example below. The first is that the probabilities of all orderings of a set of candidates sum to one. The second is that the marginal probability that one candidate precedes another has the Bradley--Terry form.

\begin{lemma}[Completions sum to one]
\label{lem:completion}
For any nonempty $R\subseteq\{1,\ldots,N\}$,
\begin{equation*}
\sum_{\tau}\ \prod_{k=1}^{|R|-1}\frac{w_{\tau_k}}{\sum_{m=k}^{|R|}w_{\tau_m}}=1,
\end{equation*}
where the sum ranges over all orderings $\tau=(\tau_1,\ldots,\tau_{|R|})$ of $R$.
\end{lemma}

\begin{proof}
By induction on $|R|$. For $|R|=1$ the product is empty and equals one. For $|R|>1$, group the orderings by their first element $i\in R$: the sum equals $\sum_{i\in R}\bigl(w_i/\sum_{j\in R}w_j\bigr)\,S(R\setminus\{i\})$, where $S(R\setminus\{i\})$ denotes the same sum over orderings of $R\setminus\{i\}$ and equals one by the induction hypothesis. The remaining factor is $\sum_{i\in R}w_i/\sum_{j\in R}w_j=1$.
\end{proof}

\begin{lemma}[Pairwise marginal]
\label{lem:pairwise}
For distinct $i$ and $j$, the total Plackett--Luce probability of the complete rankings in which $y_i$ precedes $y_j$ is $w_i/(w_i+w_j)$.
\end{lemma}

\begin{proof}
We use the standard sampling representation of the Plackett--Luce model. Draw independent exponential random variables $E_1,\ldots,E_N$ with rates $w_1,\ldots,w_N$ and rank the candidates by increasing $E_m$. The minimum is attained by candidate $m$ with probability $w_m/\sum_{m'}w_{m'}$. Conditional on the minimum time and its attaining candidate, the memoryless property makes the residual waiting times of the remaining candidates independent exponentials with their original rates. Iterating over positions gives exactly the sequential choice probabilities in \eqn{eq:pl}, so the induced distribution over rankings is the Plackett--Luce model. Under this representation, $y_i$ precedes $y_j$ if and only if $E_i<E_j$, which has probability $w_i/(w_i+w_j)$.
\end{proof}

\begin{corollary}[Special cases of the partial-order likelihood]
\label{cor:examples}
(i) If verification establishes no comparisons, then $\LE(\RV)$ contains every ranking, $p_\theta(\RV\mid x,Y)=1$, and the loss and its gradient vanish.
(ii) If $y_1$ is the only correct response among $N$ responses and $\RV=\{(1,j):j\neq 1\}$, then $p_\theta(\RV\mid x,Y)=w_1/\sum_{j=1}^{N}w_j$.
(iii) If $\RV=\{(1,3)\}$, then $p_\theta(\RV\mid x,Y)=w_1/(w_1+w_3)$, as in \eqn{eq:pareto-three}, so the loss equals the DPO loss between $y_1$ and $y_3$.
\end{corollary}

\begin{proof}
(i) is Lemma~\ref{lem:completion} with $R=\{1,\ldots,N\}$. For (ii), every compatible ranking selects $y_1$ first, which has probability $w_1/\sum_j w_j$, and the orderings of the remaining responses that follow sum to one by Lemma~\ref{lem:completion}. (iii) is Lemma~\ref{lem:pairwise} with $i=1$ and $j=3$.
\end{proof}

\paragraph{Learning comparisons jointly.}
With multiple comparisons, the likelihood accounts for their dependence through shared responses. For example, suppose $y_1$ is correct and $y_2,y_3$ are incorrect, giving $\RV=\{(1,2),(1,3)\}$. The two compatible rankings are $(1,2,3)$ and $(1,3,2)$: both place $y_1$ first, while leaving the incorrect responses unordered. Their total probability is
\begin{equation}
p_\theta(\RV\mid x,Y)
=\frac{e^{s_1}}{e^{s_1}+e^{s_2}+e^{s_3}}.
\label{eq:binary-three}
\end{equation}
Once $y_1$ is selected first, the conditional probabilities of the two remaining orderings sum to one. At equal scores, the probability that $y_1$ precedes both other responses is $1/3$. Each pairwise comparison has probability $1/2$, so multiplying the two comparison probabilities as if they were independent would instead give $1/4$. The negative log-likelihood of the joint event therefore differs from the sum of the pairwise comparison losses.

\paragraph{Example policy updates.}
In \eqn{eq:binary-three}, write $p_i=e^{s_i}/\sum_{j=1}^{3}e^{s_j}$. The coefficients are $(\Arank{1},\Arank{2},\Arank{3})=(1-p_1,-p_2,-p_3)$. At equal scores, they are $(2/3,-1/3,-1/3)$, favoring the correct response relative to both incorrect alternatives. As the model assigns greater probability to placing the correct response first, its positive coefficient decreases. Both incorrect responses receive a signal because each is compared with the correct response; no comparison between them is needed. By contrast, $y_2$ in \eqn{eq:pareto-three} has zero score derivative because it participates in no comparison. A zero score derivative does not prevent its response probability from changing through shared policy parameters.

\subsection{Computing the Partial-Order Likelihood}
\label{app:recurrence}
\label{sec:computation}

In the binary-correctness example in \eqn{eq:binary-three}, we did not need to evaluate both complete rankings separately. Once the correct response was selected first, every order of the remaining responses was acceptable, and their probabilities summed to one. The same principle gives a recurrence for general partial-order likelihoods: many compatible rankings share the same remaining candidates, so their possible continuations can be handled together.

Let $R\subseteq\{1,\ldots,N\}$ be the remaining candidate indices, write $w_i=e^{s_i}$, and let $H(R)$ be the probability of completing the ranking without violating a comparison. A candidate is eligible for the next position when none of its required predecessors remain. Denote these candidates by
\begin{equation}
\Max_{\RV}(R)
=\{i\in R:\text{there is no }j\in R
\text{ with }(j,i)\in\RV\}.
\label{eq:maximal}
\end{equation}
The likelihood can be computed recursively by adding the probabilities of the eligible next choices and their valid continuations:
\begin{equation}
H(\varnothing)=1,
\qquad
H(R)
=\sum_{i\in\Max_{\RV}(R)}
\frac{w_i}{\sum_{j\in R}w_j}
\,H(R\setminus\{i\}).
\label{eq:recurrence}
\end{equation}
The desired probability is $p_\theta(\RV\mid x,Y)=H(\{1,\ldots,N\})$. In the binary example, only $y_1$ is eligible initially, and $H(\{2,3\})=1$ after it is selected. The recurrence therefore gives $H(\{1,2,3\})=w_1/(w_1+w_2+w_3)$, as before. This also explains why the denominator must include all remaining candidates. Normalizing over the eligible set $\{1\}$ alone would always give probability one and remove the learning signal. The PL model can choose an incorrect first response; the likelihood measures the probability that it avoids doing so.

Dynamic programming stores each remaining-set value once, avoiding explicit enumeration of complete rankings. There are at most $2^N$ subset states and at most $N$ outgoing transitions per state, so the general computation can remain exponential. The number of reachable states depends on the comparison structure. Evaluating the recurrence in log space with log-sum-exp operations also allows differentiation through the policy scores without listing the compatible rankings or their conditional probabilities.

For $R\subseteq\{1,\ldots,N\}$, let $\LE_R(\RV)$ be the set of orderings of $R$ that satisfy every comparison $(i,j)\in\RV$ with $i,j\in R$, and define
\begin{equation}
H^\star(R)=\sum_{\tau\in\LE_R(\RV)}\ \prod_{k=1}^{|R|-1}\frac{w_{\tau_k}}{\sum_{m=k}^{|R|}w_{\tau_m}},
\qquad H^\star(\varnothing)=1.
\label{eq:app-hstar}
\end{equation}
By \eqn{eq:polpo}, $H^\star(\{1,\ldots,N\})=p_\theta(\RV\mid x,Y)$.

\begin{proposition}[Recurrence]
\label{prop:recurrence}
$H^\star$ satisfies the recurrence \eqn{eq:recurrence}. Hence the value $H(\{1,\ldots,N\})$ computed by \eqn{eq:recurrence} equals $p_\theta(\RV\mid x,Y)$.
\end{proposition}

\begin{proof}
Fix a nonempty $R$. An ordering $\tau$ of $R$ satisfies the comparisons within $R$ if and only if its first element has no predecessor in $R$, that is, $\tau_1\in\Max_{\RV}(R)$, and the remaining ordering $(\tau_2,\ldots,\tau_{|R|})$ satisfies the comparisons within $R\setminus\{\tau_1\}$. Grouping the orderings in \eqn{eq:app-hstar} by their first element therefore gives
\begin{equation*}
H^\star(R)=\sum_{i\in\Max_{\RV}(R)}\frac{w_i}{\sum_{j\in R}w_j}\,H^\star(R\setminus\{i\}),
\end{equation*}
which is \eqn{eq:recurrence}. Both $H$ and $H^\star$ equal one on the empty set and satisfy the same recurrence, so they agree on every subset by induction on $|R|$.
\end{proof}

The subsets reachable from $\{1,\ldots,N\}$ by repeatedly removing eligible candidates are exactly the subsets $R$ that are closed under the relation in the following sense: if $i\in R$ and $(i,j)\in\RV$, then $j\in R$. These subsets are the order ideals under the convention that a preferred candidate is greater, and the eligible candidates of an ideal are its maximal elements. A top-down implementation that memoizes reachable states and expands every nonempty state therefore visits one state per order ideal, including the empty set. This count equals $2^N$ only when $\RV$ is empty, equals $N+1$ when $\RV$ is a complete ranking, and equals $1+\sum_{g=1}^{M}(2^{|G_g|}-1)$ for an ordered partition $G_1\succ\cdots\succ G_M$. Early termination when no comparisons remain can reduce the number of evaluated states. These are counts of reachable recurrence states. The batched evaluation instead evaluates all subsets within each series component, grouped by cardinality (see Appendix~\ref{app:general-orders}).

\subsection{Ordered Partitions}
\label{app:blocks}

Let $\RV=G_1\succ\cdots\succ G_M$ be an ordered partition of $\{1,\ldots,N\}$: every candidate in an earlier block precedes every candidate in a later block, and no ordering is specified within a block. Binary correctness and graded scalar feedback induce this structure. General Pareto relations need not do so. Write $W_{>g}=\sum_{h>g}\sum_{j\in G_h}w_j$ for the total weight of the blocks below $G_g$, with $W_{>M}=0$. For $S\subseteq G_g$, define
\begin{equation}
F_g(S)=\sum_{\tau}\ \prod_{k=1}^{|S|}\frac{w_{\tau_k}}{\sum_{m=k}^{|S|}w_{\tau_m}+W_{>g}},
\qquad F_g(\varnothing)=1,
\label{eq:app-block-factor}
\end{equation}
where the sum ranges over all orderings $\tau$ of $S$. Thus $F_g(G_g)$ is the probability that the sequential choice process selects every candidate in $G_g$ before any candidate in a lower block, given that all higher blocks have already been selected.

\begin{proposition}[Block factorization]
\label{prop:blocks}
The likelihood factorizes as
\begin{equation}
p_\theta(\RV\mid x,Y)=\prod_{g=1}^{M-1}F_g(G_g).
\label{eq:block-product}
\end{equation}
Each factor satisfies
\begin{equation}
F_g(S)=\sum_{i\in S}\frac{w_i}{\sum_{j\in S}w_j+W_{>g}}\,F_g(S\setminus\{i\}).
\label{eq:app-block-recurrence}
\end{equation}
Once the block weights $W_{>g}$ are computed, evaluating all factors by this recurrence takes $O\bigl(\sum_{g=1}^{M-1}|G_g|\,2^{|G_g|}\bigr)$ arithmetic operations.
\end{proposition}

\begin{proof}
The linear extensions of an ordered partition are the concatenations $\sigma=(\tau^{(1)},\ldots,\tau^{(M)})$ of arbitrary orderings $\tau^{(g)}$ of the blocks $G_g$. When position $k$ of $\sigma$ lies in block $g$, the remaining candidates are the unselected members of $G_g$ together with all lower blocks, so the denominator of the choice probability in \eqn{eq:pl} is the weight of the unselected members of $G_g$ plus $W_{>g}$. The product over positions therefore factorizes over blocks, and summing over the orderings of each block independently gives $\prod_{g=1}^{M}F_g(G_g)$. For the final block, $W_{>M}=0$ and $F_M(G_M)=1$ by Lemma~\ref{lem:completion}. The recurrence follows by grouping the orderings of $S$ by their first element, as in Proposition~\ref{prop:recurrence}. The cost follows because block $g$ has $2^{|G_g|}$ subset states with at most $|G_g|$ transitions each.
\end{proof}

\paragraph{Extension to series components.}
The same factorization applies when the ordered blocks retain internal comparisons. Form the undirected incomparability graph of $\RV$, with an edge between distinct candidates $i$ and $j$ exactly when neither $(i,j)$ nor $(j,i)$ belongs to $\RV$. Its connected components can be ordered as $C_1\succ\cdots\succ C_M$, meaning that every candidate in $C_g$ precedes every candidate in $C_h$ whenever $g<h$. Unlike tie blocks, a component need not be an antichain: candidates linked by a path of incomparable pairs may still be comparable to one another. The original comparisons within each component must therefore be retained. For this decomposition, write $W_{>g}=\sum_{h>g}\sum_{j\in C_h}w_j$ and let $\RV|_{C_g}$ be the restriction of $\RV$ to $C_g$.

\begin{proposition}[Series-component factorization]
\label{prop:series-components}
For $S\subseteq C_g$, start with $S$ and all lower components as the remaining candidates. Let $Q_g(S)$ be the probability of selecting all candidates in $S$, in an order satisfying every comparison in $\RV$ whose endpoints both lie in $S$, before selecting any candidate in a lower component. Then
\begin{equation}
Q_g(\varnothing)=1,
\qquad
Q_g(S)=\sum_{i\in\Max_{\RV|_{C_g}}(S)}
\frac{w_i}{\sum_{j\in S}w_j+W_{>g}}\,Q_g(S\setminus\{i\}),
\label{eq:series-component-recurrence}
\end{equation}
and
\begin{equation}
p_\theta(\RV\mid x,Y)=\prod_{g=1}^{M}Q_g(C_g).
\label{eq:series-component-product}
\end{equation}
For an ordered partition, the components are the tie blocks and $Q_g=F_g$. The final factor is one if the final component is an antichain. If it contains an internal comparison, its factor must be evaluated even though $W_{>M}=0$.
\end{proposition}

\begin{proof}
Candidates in distinct incomparability components are comparable. Their comparison direction is uniform across the two components: if $a$ precedes $b$ and $a'$ is incomparable with $a$ in the same component, then $b$ cannot precede $a'$, since transitivity would imply that $a$ precedes $a'$. Thus $a'$ precedes $b$. Propagating along incomparability paths in both components gives the same direction for every cross-component pair. Transitivity then orders the components totally.

Every compatible full ranking is consequently a concatenation of compatible internal rankings of $C_1,\ldots,C_M$. Within component $g$, the remaining candidates consist of its unselected members and all lower components. Grouping internal rankings by their first candidate gives \eqn{eq:series-component-recurrence}; its eligible set enforces all internal comparisons. Factoring each full-ranking probability over components and summing separately over their compatible internal rankings gives \eqn{eq:series-component-product}, without assuming independence of the comparison events. In a final antichain, all internal rankings are allowed and their probabilities sum to one by Lemma~\ref{lem:completion}. If the final component contains a comparison, some internal rankings are forbidden and have positive PL probability because all weights are positive, so its factor is strictly less than one.
\end{proof}

\section{Likelihood Computation and Implementation Details}
\label{app:implementation}

This appendix describes how we evaluate and optimize the POLPO loss in \eqn{eq:polpo}. Scalar verifier scores induce an ordered partition of tied responses, whereas the Pareto rule in \eqn{eq:pareto} preserves the comparisons supported by individual verification outcomes. Both cases use the same partial-order likelihood.

\subsection{POLPO Training Algorithm}
\label{app:training-algorithm}
\label{sec:training-objective}

POLPO alternates between collecting responses and updating the policy (see Algorithm~\ref{alg:polpo}). At each rollout iteration, a frozen behavior policy $\pi_{\mathrm{old}}$ generates $N$ responses for each of $B_{\mathrm{roll}}$ prompts. Verification determines the comparison relation within each response group. The collected groups are used for $E$ training epochs, with the group loss defined in \eqn{eq:polpo}.

The reference policy $\pi_{\mathrm{ref}}$ remains fixed throughout training. Sampled responses, verifier outcomes, comparison relations, and reference log probabilities are treated as constants during policy updates. Current-policy scores are recomputed at each update, and gradients flow through these scores and the likelihood calculation. Sampling, verification, and group selection are not differentiated.

\begin{algorithm}[t]
\caption{Partial-Order Likelihood Policy Optimization (POLPO)}
\label{alg:polpo}
\small
\noindent\textbf{Input:} Initial policy $\pi_\theta$, fixed reference $\pi_{\mathrm{ref}}$, prompt distribution $\rho$, verification function $\mathbf c$, comparison rule $\mathcal A$ (Pareto dominance by default), and score scale $\beta>0$; group size $N\geq2$, rollout batch size $B_{\mathrm{roll}}$, training epochs per rollout $E$, rollout iterations $T$, and an optimizer.
\par\smallskip
\begin{tabular}{@{}r@{\quad}p{0.90\linewidth}@{}}
1 & \textbf{for} $t=1,\ldots,T$ \textbf{do} \\
2 & \hspace*{1em}$\pi_{\mathrm{old}}\gets\pi_\theta$; hold this rollout policy fixed \\
3 & \hspace*{1em}Sample prompts $x_1,\ldots,x_{B_{\mathrm{roll}}}\sim\rho$ \\
4 & \hspace*{1em}\textbf{for} $b=1,\ldots,B_{\mathrm{roll}}$ \textbf{do} \\
5 & \hspace*{2em}Sample $Y_b=(y_{b1},\ldots,y_{bN})$ from $\pi_{\mathrm{old}}(\cdot\mid x_b)$ \\
6 & \hspace*{2em}Evaluate $\mathbf c_{bi}\gets\mathbf c(x_b,y_{bi})$ for $i=1,\ldots,N$ \\
7 & \hspace*{2em}$\mathcal R_{V,b}\gets\mathcal A(\mathbf c_{b1},\ldots,\mathbf c_{bN})$ \\
8 & \hspace*{1em}\textbf{end for} \\
9 & \hspace*{1em}Prepare training groups using the inclusion conditions (see Appendix~\ref{app:likelihood-conditions}) \\
10 & \hspace*{1em}\textbf{for} $e=1,\ldots,E$ \textbf{do} \\
11 & \hspace*{2em}\textbf{for each} training batch $\mathcal D$ of response groups \textbf{do} \\
12 & \hspace*{3em}Evaluate $\ell_{\mathrm{POLPO}}(\theta;x,Y,\mathcal R_V)$ from \eqn{eq:polpo} for each $(x,Y,\mathcal R_V)\in\mathcal D$ \\
13 & \hspace*{3em}Update $\theta$ with the optimizer using the gradients of these group losses \\
14 & \hspace*{2em}\textbf{end for} \\
15 & \hspace*{1em}\textbf{end for} \\
16 & \textbf{end for} \\
17 & \textbf{return} $\pi_\theta$ \\
\end{tabular}
\end{algorithm}

\paragraph{Likelihood evaluation.}
The group loss in \eqn{eq:polpo} is evaluated using the series-component factorization in Proposition~\ref{prop:series-components}, with each required factor computed by a subset recurrence in log space. Policy scores are recomputed from the current parameters according to \eqn{eq:dpo-score}. At every step of the recurrence, the choice denominator includes all remaining candidates, including candidates that cannot be selected next without violating the relation. The computation is differentiable through the policy scores and does not enumerate complete rankings. An empty comparison relation has log-likelihood zero.

\subsection{Log-Space Block Recurrence}
\label{app:log-space}

For ordered partitions $G_1\succ\cdots\succ G_M$, the evaluator uses the factors in Proposition~\ref{prop:blocks}. Equal scalar verifier scores form a tie block; binary correctness yields at most two blocks. To evaluate the factors stably, let $\bar{s}_{>g}=\log W_{>g}$ and $\phi_g(S)=\log F_g(S)$. For a non-final block, $\bar{s}_{>g}$ is the log-sum-exp of the scores of all lower candidates. With $\phi_g(\varnothing)=0$, the log-space recurrence is
\begin{equation}
\phi_g(S)
=\lse_{i\in S}
\Bigl[
s_i-\lse\bigl((s_j)_{j\in S},\bar{s}_{>g}\bigr)
+\phi_g(S\setminus\{i\})
\Bigr].
\label{eq:app-log-recurrence}
\end{equation}
The inner log-sum-exp includes one entry for every remaining candidate and one entry for the aggregate weight of the lower blocks; repeated score values retain their multiplicities. The final tie block contributes a factor of one and is omitted, giving $\log p_\theta(\RV\mid x,Y)=\sum_{g=1}^{M-1}\phi_g(G_g)$.

The implementation enumerates all subsets of a processed block as bit masks, groups them by cardinality, and caches score-independent index and membership tables for each block size and device. All subsets of a given cardinality are evaluated together by tensor gathers and log-sum-exp operations. Processing cardinalities in increasing order ensures that the values for $S\setminus\{i\}$ are available when computing the value for $S$.

Automatic differentiation through this recurrence and the sequence-summed scores yields the coefficients in Proposition~\ref{prop:po-gradient}. Differentiating the log-likelihood gives $\Arank{i}$; differentiating its negative gives $-\Arank{i}$. Neither linear extensions nor their conditional probabilities are explicitly constructed. Reference log probabilities are computed for the sampled responses and treated as constants.

\subsection{General Partial Orders}
\label{app:general-orders}

For binary multi-verifier feedback, Pareto dominance in \eqn{eq:pareto} is equivalent to strict inclusion between sets of satisfied constraints. These sets are represented as integer bit masks for efficient comparison. Every response used to construct the relation must have recorded verifier outcomes; missing outcomes are not interpreted as failed checks.

The evaluator constructs the incomparability graph of the retained candidates and finds its connected components $C_1,\ldots,C_M$. As shown in Proposition~\ref{prop:series-components}, these components can be ordered so that every candidate in an earlier component dominates every candidate in a later one. Internal comparisons within a component are retained. This decomposition does not impose additional comparisons between incomparable responses.

Each component factor $Q_g$ is evaluated using \eqn{eq:series-component-recurrence}. In log space, the update has the form of \eqn{eq:app-log-recurrence}, with the outer sum restricted to $\Max_{\RV}(S)$ and the denominator unchanged. If there are no lower components, the lower-weight term is omitted from the denominator. A final component is omitted only when it contains no internal comparison; a final component with internal comparisons still requires evaluation. For example, if the only comparison is $y_1\succ y_3$ and $y_2$ is incomparable with both, all three candidates belong to one component. Its factor is evaluated and equals $w_1/(w_1+w_3)$.

\paragraph{Vectorized evaluation and computational cost.}
The vectorized computation evaluates all $2^m$ subsets of each processed component of size $m$, rather than traversing only the reachable order ideals (see Appendix~\ref{app:recurrence}). For each subset, candidates with a predecessor still present are excluded from the eligible set. Values for unreachable subsets are computed but do not contribute to the final component likelihood. The denominator retains all candidates in the subset and all lower components.

For a component without internal comparisons, the subset transitions require $O(m2^m)$ arithmetic operations after the lower-component weight is available. For a component with internal comparisons, determining the eligible candidates uses a dense matrix multiplication between subset membership indicators and the dominance matrix; this requires $O(m^2 2^m)$ arithmetic operations. The cached index and membership tables contain $O(m2^m)$ entries. These implementation costs differ from counting only the reachable states of a top-down recurrence. Appendix~\ref{app:cost} reports measured times.

\paragraph{Evaluation limits and candidate filtering.}
\label{app:likelihood-conditions}
Groups with missing or empty responses are excluded before any optional filtering. An optional filter removes responses that reach the generation-length limit; the likelihood is then evaluated on the remaining responses and their induced partial order. Groups with fewer than two responses or no strict comparison contribute zero loss. In the reported training runs, each series component requiring dynamic programming is limited to $16$ candidates, and groups exceeding this limit are omitted; the fixed-block cost experiment below varies this limit. A final antichain requires no computation and is exempt. Consequently, the component-size limit cannot exclude a group when $N\leq16$.

\subsection{Computational Cost}
\label{app:cost}

We measured the cost of the exact likelihood on NVIDIA B300 GPUs in the four coding settings of Section~\ref{sec:exp-coding}.
To isolate the likelihood, we replayed five training updates per setting and timed the forward and backward computation of the loss for the response groups that one GPU processes in an update.
The timing starts from given token log probabilities, because the forward and backward passes of the policy are the same for POLPO and GRPO.
At the group size $N=8$ of these runs, the POLPO loss, including the construction of the Pareto relation and the evaluation of the likelihood, takes 26--203\,ms longer per update than GRPO's clipped objective (Table~\ref{tab:cost}).
This overhead amounts to 0.03--0.07\% of the time of a training update, and stays below 0.7\% even relative to the policy update alone, which takes 8--31\,s.

\begin{figure}[t]
\centering
\includegraphics[width=0.65\linewidth]{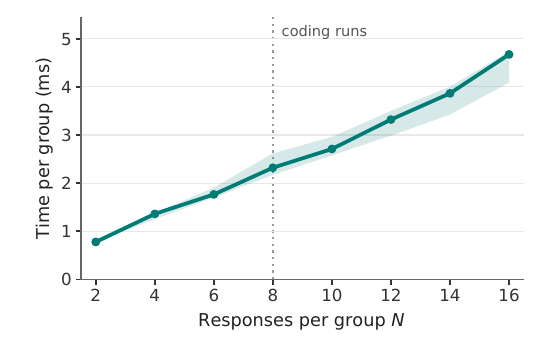}
\caption{Time for the likelihood and its gradient per response group on one GPU, as a function of the group size $N$.
Only training groups with at least one comparison are included.
The line shows the median over the four coding settings in Table~\ref{tab:cost}, and the band their range; groups larger than eight are available only for MBPP and APPS.
The dotted line marks the group size of the coding runs.
Peak additional GPU memory, including cached index tables, is below 0.1\,MiB at $N=8$ and below 33\,MiB at $N=16$.}
\label{fig:likelihood-group-size}
\end{figure}

Figure~\ref{fig:likelihood-group-size} shows how the time per group depends on the group size $N$.
We use the Pareto relations of training groups that contain at least one comparison.
Groups with more than eight responses combine responses to the same problem from different updates, which is possible for MBPP and APPS.
The time grows from about 0.8\,ms at $N=2$ to 2.2--2.6\,ms at $N=8$ and 4.1--4.7\,ms at $N=16$, the group size of the IFM5 and MulDimIF runs.
To measure the overhead at $N=16$ in training, we also ran POLPO with 16 responses per prompt for eleven updates in each coding setting, otherwise as in the runs above, and replayed their second to eleventh updates in the same way.
The overhead over GRPO's loss was then 69--417\,ms per update, or 0.04--0.08\% of a training update (Table~\ref{tab:cost}).

\begin{table}[t]
\centering
\small
\caption{Overhead of the POLPO loss over GRPO's loss in the coding experiments (medians), at the group size of the coding runs ($N=8$) and at $N=16$.
\emph{Training update} is the time of one training update in the POLPO runs, from the training logs.
\emph{Loss} is the time to compute each loss and its gradient for the response groups that one GPU processes in an update, measured by replaying five training updates at $N=8$ and ten at $N=16$.
\emph{Overhead} is the difference between the two loss times, in milliseconds and as a percentage of the training-update time.}
\label{tab:cost}
\begin{tabular}{lrrrrr}
\toprule
& Training & \multicolumn{2}{c}{Loss (ms)} & \multicolumn{2}{c}{Overhead} \\
\cmidrule(lr){3-4}\cmidrule(lr){5-6}
Setting & update (s) & POLPO & GRPO & (ms) & (\%) \\
\midrule
\multicolumn{6}{l}{\emph{$N=8$}} \\
MBPP, Qwen3-0.6B & 57.5 & 64.2 & 24.2 & 40.0 & 0.07 \\
MBPP, Qwen3-4B & 85.3 & 49.9 & 24.4 & 25.5 & 0.03 \\
APPS Introductory & 155.2 & 72.6 & 23.9 & 48.7 & 0.03 \\
AceCode & 375.1 & 298.3 & 95.4 & 202.9 & 0.05 \\
\midrule
\multicolumn{6}{l}{\emph{$N=16$}} \\
MBPP, Qwen3-0.6B & 110.6 & 102.5 & 24.2 & 78.3 & 0.07 \\
MBPP, Qwen3-4B & 170.8 & 93.9 & 24.7 & 69.2 & 0.04 \\
APPS Introductory & 235.8 & 143.0 & 24.7 & 118.3 & 0.05 \\
AceCode & 548.2 & 514.7 & 97.8 & 416.9 & 0.08 \\
\bottomrule
\end{tabular}
\end{table}

We also report loss-computation costs for initial-policy instruction-following rollouts in Table~\ref{tab:if-cost}.
Relative to GRPO, the POLPO loss takes 24.5\,ms longer per update on IFM5, 74.4\,ms longer on MulDimIF, and 22.7\,ms longer on MathIF with thinking enabled.
These differences correspond to 0.05\%, 0.12\%, and 0.03\% of the reported training-update times, respectively.
The percentages compare the loss-time differences with the listed update times; they are not direct measurements of differences in total training time.

\begin{table}[t]
\centering
\small
\setlength{\tabcolsep}{5pt}
\caption{POLPO loss overhead relative to GRPO in instruction-following settings using initial-policy rollouts. Loss times are per update; overhead is the difference between the POLPO and GRPO loss times, in milliseconds and as a percentage of the listed training-update time.}
\label{tab:if-cost}
\begin{tabular}{lrrrrr}
\toprule
& Training & \multicolumn{2}{c}{Loss (ms)} & \multicolumn{2}{c}{Overhead} \\
\cmidrule(lr){3-4}\cmidrule(lr){5-6}
Setting & update (s) & POLPO & GRPO & (ms) & (\%) \\
\midrule
IFM5, Qwen3-4B & 50.7 & 43.6 & 19.1 & 24.5 & 0.05 \\
MulDimIF, Qwen3-4B & 60.6 & 93.9 & 19.5 & 74.4 & 0.12 \\
MathIF thinking, Qwen3-0.6B & 75.8 & 42.2 & 19.5 & 22.7 & 0.03 \\
\bottomrule
\end{tabular}
\end{table}

\subsubsection{Update Cost and Block-Size Stress Test}
\label{app:if-update-cost}

We also measure update time and GPU memory on IFM5 rollouts from a 0.6B model, separately from the isolated loss timings above. The benchmark uses one 96\,GB NVIDIA GH200 GPU, 256 sequences per step, and one response group per microbatch for both POLPO and GRPO. Table~\ref{tab:if-update-cost} reports means over steps 2--4. At fixed sequence count, increasing $N$ decreases the number of groups per step. This 0.6B benchmark has different conditions from the 4B IFM5 experiment in Table~\ref{tab:if-cost}.

\begin{table}[t]
\centering
\small
\setlength{\tabcolsep}{4pt}
\caption{Update cost on IFM5 0.6B rollouts, averaged over steps 2--4 on one GH200. Each step contains 256 sequences, with one group per microbatch. Times, memory, and the reported loss time are shown for the measured updates.}
\label{tab:if-update-cost}
\begin{tabular}{rrrrrrr}
\toprule
& \multicolumn{2}{c}{Update (s)} & & \multicolumn{2}{c}{GPU memory (GB)} & Loss \\
\cmidrule(lr){2-3}\cmidrule(lr){5-6}
$N$ & GRPO & POLPO & Ratio & GRPO & POLPO & (s/step) \\
\midrule
8  & 10.82 & 10.79 & 1.00$\times$ & 16.8 & 19.1 & 0.24 \\
16 & 5.77  & 6.16  & 1.07$\times$ & 19.5 & 18.3 & 0.23 \\
\bottomrule
\end{tabular}
\end{table}

At $N=16$, POLPO takes 6.16\,s per update on these rollouts, compared with 5.77\,s for GRPO (1.07$\times$), and uses 18.3\,GB of GPU memory compared with 19.5\,GB for GRPO.

To test a demanding comparison structure, we replace the relation in every group with two ordered tiers: $m$ tied responses above the remaining $N-m$ responses. Here $m$ is the size of the upper component, consistent with the component-size notation in Appendix~\ref{app:general-orders}. Table~\ref{tab:if-block-cost} reports POLPO update time and memory, the ratio to the GRPO update time at the corresponding $N$ in Table~\ref{tab:if-update-cost}, and the loss time per group. Likelihoods are evaluated for every group in these fixed-block runs, with no exclusions or identical-outcome groups. The constructed relation is a controlled stress test rather than a measurement of the natural distribution of orders.

\begin{table}[t]
\centering
\small
\setlength{\tabcolsep}{5pt}
\caption{Fixed-block cost on one GH200: each group has $m$ tied upper responses and $N-m$ lower responses. The likelihood is evaluated for all groups. Update time, GPU memory, and per-group loss time are averaged over steps 2--4; the ratio uses the GRPO update time at the same $N$ in Table~\ref{tab:if-update-cost}.}
\label{tab:if-block-cost}
\begin{tabular}{rrrrrr}
\toprule
$N$ & $m$ & Update (s) & GRPO ratio & Memory (GB) & Loss (ms/group) \\
\midrule
8  & 4  & 10.96  & 1.01$\times$ & 19.5 & 9.6 \\
8  & 7  & 10.89  & 1.01$\times$ & 20.4 & 13.5 \\
\midrule
16 & 4  & 6.23   & 1.08$\times$ & 19.0 & 9.5 \\
16 & 8  & 5.93   & 1.03$\times$ & 17.8 & 14.3 \\
16 & 12 & 6.51   & 1.13$\times$ & 17.7 & 21.0 \\
16 & 15 & 6.52   & 1.13$\times$ & 18.3 & 25.8 \\
\bottomrule
\end{tabular}
\end{table}

In the fixed-block stress test at $N=16$ and $m=15$, POLPO takes 6.52\,s per update (1.13$\times$ GRPO) and uses 18.3\,GB of GPU memory. Together with the real-rollout result, these measurements suggest modest time and memory costs at $N=16$ in the tested settings, including this demanding fixed-block case. The constructed two-tier setting does not cover every possible partial order.

\FloatBarrier
\section{Experimental Details}
\label{app:experimental-details}

This appendix describes the datasets, training configurations, and evaluation protocols used in the experiments.

\subsection{Controlled Program Synthesis}
\label{app:toy-details}

Each program is specified by three independent categorical choices $(d_j,v_j)$, one for each branch $j\in\{1,2,3\}$.
Each choice has $3(M-1)$ possible values, with $d_j\in\{2,\ldots,M\}$ and $v_j\in\{\texttt{Fizz},\texttt{Buzz},\texttt{FizzBuzz}\}$.
The program executes the first satisfied branch:
\begin{verbatim}
def f(n):
    if n % d1 == 0:
        return v1
    elif n % d2 == 0:
        return v2
    elif n % d3 == 0:
        return v3
    return str(n)
\end{verbatim}
Branch distributions are initialized uniformly, and duplicate choices are allowed.
The log probability of a program is the sum of the log probabilities of its three branch choices.
For target divisors $(a,b)$, the intended output is \texttt{FizzBuzz} for a multiple of both divisors, \texttt{Fizz} or \texttt{Buzz} for a multiple of only one, and the input string otherwise.
For the divisors and search spaces considered here, the two completely correct programs begin with $(ab,\texttt{FizzBuzz})$ and place $(a,\texttt{Fizz})$ and $(b,\texttt{Buzz})$ in either order afterward.
Their initial total probability is $2/[3(M-1)]^3$.

Each integer input from $1$ to $60$ defines one verifier.
The first sweep uses $(a,b)=(3,5)$ and $M\in\{15,30,60\}$; the second uses $M=60$ and $(a,b)\in\{(3,5),(3,7),(3,11)\}$.
For these three divisor pairs, respectively, 4, 2, and 1 of the 60 inputs are divisible by both divisors.
The other input-category counts also change in the second sweep.
The experiment therefore varies the specification and does not attribute differences solely to the frequency of overlap inputs.

Each condition uses five training seeds (42, 43, 44, 45, and 46), 1,000 updates, and 64 independent groups of eight programs per update, or 512 new programs per update.
Each group contains responses to the same synthesis problem; groups are independent draws from the current policy.
Optimization uses Adam with learning rate $0.03$, moment coefficients $(0.9,0.999)$, and $\epsilon=10^{-8}$.
The GRPO analogue uses rewards from \eqn{eq:reward-baselines}, centered and divided by the population standard deviation within each group, with no KL penalty.
POLPO uses $\beta=0.1$, the fixed initial uniform policy as reference, and the group loss in \eqn{eq:polpo}.
The verifier relation is built from the complete 60-dimensional outcome vector.
POLPO--Rate uses the same policy scores, $\beta$, reference, group-loss normalization, and optimizer, but replaces this relation by $y_i\succ y_j$ whenever $\sum_{n=1}^{60}c_n(y_i)>\sum_{n=1}^{60}c_n(y_j)$.
Since all programs are evaluated on the same 60 inputs, this is exactly the ranking by mean test pass rate.
Programs with equal rates remain unordered, and the likelihood marginalizes over their relative order.
POLPO--All instead sets $y_i\succ y_j$ whenever $y_i$ passes all 60 inputs and $y_j$ does not, leaving all other pairs unordered.
Since a completely correct program dominates every other program, these are exactly the comparisons of POLPO in which the preferred program is completely correct.

Complete-success probability is evaluated exactly by summing the probabilities of the two correct programs after every update.
The first update at which it reaches $0.9$ defines the sample-efficiency statistic, multiplied by 512 to obtain the number of generated programs.
No interpolation between recorded updates is needed.
Input-specific success probabilities are also computed from the categorical distributions, so evaluation has no response-sampling noise; the training trajectory still depends on the seed.
Learning curves report the pointwise mean and sample standard deviation across the five training seeds, without smoothing.
Threshold-crossing budgets are computed separately for each seed; when any run fails to reach the threshold within the training budget, we report the number of runs that reach it and do not compute a mean from only the successful runs.

\subsection{Coding Tasks}
\label{app:coding-details}

\subsubsection{Datasets and Prompts}

\paragraph{MBPP.}
We use the \texttt{full} configuration of \texttt{google-research-datasets/mbpp}, with the official 374-problem training split and 500-problem test split \citep{austin2021mbpp}.
Each problem has three test assertions.
The prompt requests a self-contained Python implementation in a Markdown code block and includes the problem description and the first test as an example.
All three assertions, including the displayed example, are used to verify the response.
The released reference implementations are not used for supervised warmup or added to response groups.
The 90-problem validation split is separate from the reported 500-problem test evaluation; results use the prescribed final checkpoint rather than a checkpoint selected by validation or test performance.

\paragraph{APPS Introductory.}
We use the Introductory tier of \texttt{codeparrot/apps} \citep{hendrycks2021apps}.
Of its 2,639 training problems, 241 have no executable test cases and are excluded, leaving 2,398 training problems.
All 1,000 Introductory test problems are retained.
There is no further filtering by test count, and every available test is used.
Prompts contain the problem statement and supplied starter code when present, together with instructions appropriate for standard-input/output or function-call evaluation.
Test cases used by the checker are not appended to the prompt; examples already present in the problem statement are retained.
The training and test problem statements have no exact overlap after whitespace normalization; this check does not establish the absence of semantic near-duplicates.

\paragraph{AceCode.}
We use the hard RL-training subset released with AceCoder \citep{zeng2025acecoder}, distributed as \nolinkurl{CodeDPO/rlhf_dataset_20250126_openrlhf_format_hard} at revision \nolinkurl{f40aa9de97d882c6d411a8ca5e674ab6b1da2c12}.
All 21,262 released training rows are used.
The hard-subset selection is inherited from the release, rather than recomputed using our initial model.
Each problem supplies a prompt and a list of test-code entries.
Each entry defines one verifier and can contain setup code as well as assertions; entries execute in their supplied order within the generated program's namespace.
Thus, a verifier corresponds to a supplied test entry, not necessarily a single assertion.
The checker records all entries' outcomes, including those after a failure.
We use neither teacher responses nor a supervised warmup stage.
Because this release has no official test split, evaluation uses the external benchmarks described below.

\subsubsection{Models and Optimization}

MBPP initializes either \texttt{Qwen/Qwen3-0.6B} or \texttt{Qwen/Qwen3-4B}, using each model's native chat template with thinking enabled during both training and evaluation.
APPS and AceCode initialize \nolinkurl{Qwen/Qwen2.5-Coder-1.5B-Instruct}, also using its native chat template.
The same initial weights are used across methods within each dataset and model size. MBPP trained-model results aggregate three training seeds; APPS Introductory and AceCode use seed 42 across methods.
All experiments perform full-parameter fine-tuning.
Table~\ref{tab:coding-hyperparameters} summarizes the training hyperparameters.

\begin{table}[t]
\centering
\small
\caption{Training hyperparameters for coding tasks. Rollout and update mini-batch sizes count prompts in a full batch; the final incomplete prompt batch is retained. Learning-rate settings are listed in Table~\ref{tab:coding-learning-rates}. Token limits and sampling settings apply to training rollouts; evaluation context limits are specified in the text.}
\label{tab:coding-hyperparameters}
\begin{tabular}{lccc}
\toprule
Setting & MBPP & APPS Introductory & AceCode \\
\midrule
Initial model & Qwen3-0.6B / Qwen3-4B & \multicolumn{2}{c}{Qwen2.5-Coder-1.5B-Instruct} \\
Optimizer & AdamW & AdamW & AdamW \\
Learning rate & \multicolumn{3}{c}{See Table~\ref{tab:coding-learning-rates}} \\
Adam coefficients $(\beta_1,\beta_2)$ & $(0.9,0.999)$ & $(0.9,0.999)$ & $(0.9,0.999)$ \\
Weight decay & $0.01$ & $0.01$ & $0.01$ \\
Learning-rate schedule & Constant & Constant & Constant \\
Warmup & None & None & None \\
Gradient clipping norm & $1.0$ & $1.0$ & $1.0$ \\
Prompts per rollout batch & 32 & 64 & 256 \\
Responses per prompt & 8 & 8 & 8 \\
Update mini-batch size & 32 & 64 & 256 \\
Update epochs per rollout batch & 1 & 1 & 1 \\
Training epochs & 5 & 2 & 1 \\
Maximum prompt length (tokens) & 1,024 & 16,384 & 16,384 \\
Maximum response length (tokens) & 2,048 & 2,048 & 2,048 \\
Sampling temperature & $1.0$ & $1.0$ & $1.0$ \\
Top-$p$ & $1.0$ & $1.0$ & $1.0$ \\
Top-$k$ & Disabled & Disabled & Disabled \\
\bottomrule
\end{tabular}
\end{table}

Each training round samples eight responses per prompt at temperature $1$, with top-$p$ and top-$k$ truncation disabled.
All sampled responses contribute to a single optimizer update; the optimizer mini-batch equals the full rollout batch.
The prompt order is shuffled, and the final incomplete prompt batch in each epoch is retained.
Training uses only fresh responses, without replay or demonstrations.
Within each MBPP model size, all runs train from the common initial model for all five epochs.

The optimizer is AdamW with a constant learning rate, no warmup, moment coefficients $(0.9,0.999)$, $\epsilon=10^{-8}$, weight decay $0.01$, and gradient clipping at norm $1$.
Table~\ref{tab:coding-learning-rates} lists the learning-rate settings for each dataset, initial model, and method.

The validation-based learning-rate selection protocol specifies a random problem-level split of each eligible training set into 80\% training and 20\% validation, using seed 42 and rounding the validation count up. This gives training/validation counts of 299/75 for MBPP, 1,918/480 for APPS Introductory, and 17,009/4,253 for AceCode. All tests and responses associated with a problem remain in the same partition, and the partition is shared across methods, model sizes, and learning-rate candidates. The official test sets and external evaluation benchmarks remain separate.

The protocol compares learning rates $\{5\times10^{-7},1\times10^{-6},2\times10^{-6},5\times10^{-6},10^{-5}\}$, training each candidate on the training portion for the dataset-specific number of epochs in Table~\ref{tab:coding-hyperparameters}. For each method and initial model, selection uses the highest validation All-pass score at the final checkpoint, with one greedy response per problem. Ties are resolved by higher Avg.\ pass and then by the smaller learning rate. Training then restarts from the initial model with the selected rate on the full training set, including the validation portion, for the same number of epochs before final-checkpoint evaluation.
For the MBPP POLPO--Rate ablation, we reuse the learning rate selected for POLPO for the same initial model.
The MBPP POLPO--All and GRPO--Pareto ablations use the learning rate of GRPO--All and GRPO--Rate for the same initial model.

\begin{table}[t]
\centering
\small
\caption{Learning-rate settings for the coding experiments. A dash indicates that the method is not evaluated in that setting.}
\label{tab:coding-learning-rates}
\begin{tabular}{lcccc}
\toprule
& \multicolumn{2}{c}{MBPP} & APPS Introductory & AceCode \\
\cmidrule(lr){2-3}
Method & Qwen3-0.6B & Qwen3-4B & \multicolumn{2}{c}{Qwen2.5-Coder-1.5B-Instruct} \\
\midrule
GRPO--All & $2\times10^{-6}$ & $1\times10^{-6}$ & $2\times10^{-6}$ & $5\times10^{-7}$ \\
GRPO--Rate & $2\times10^{-6}$ & $1\times10^{-6}$ & $2\times10^{-6}$ & $5\times10^{-7}$ \\
GRPO--Pareto & $2\times10^{-6}$ & $1\times10^{-6}$ & -- & -- \\
REINFORCE++--All & $2\times10^{-6}$ & $2\times10^{-6}$ & $2\times10^{-6}$ & -- \\
REINFORCE++--Rate & $2\times10^{-6}$ & $2\times10^{-6}$ & $2\times10^{-6}$ & -- \\
GDPO & $10^{-6}$ & $2\times10^{-6}$ & $2\times10^{-6}$ & -- \\
GraphDPO & $2\times10^{-6}$ & $5\times10^{-6}$ & $2\times10^{-6}$ & $5\times10^{-7}$ \\
DPO--Pareto & $2\times10^{-6}$ & $1\times10^{-6}$ & $2\times10^{-6}$ & $5\times10^{-7}$ \\
POLPO--All & $2\times10^{-6}$ & $1\times10^{-6}$ & -- & -- \\
POLPO--Rate & $5\times10^{-6}$ & $5\times10^{-6}$ & -- & -- \\
POLPO & $5\times10^{-6}$ & $5\times10^{-6}$ & $2\times10^{-6}$ & $2\times10^{-6}$ \\
\bottomrule
\end{tabular}
\end{table}

GRPO--All and GRPO--Rate use clipping parameter $0.2$, standardize rewards within each response group, and average token contributions within each response before averaging over responses.
Neither baseline uses a KL or entropy penalty.

POLPO uses $\beta=10^{-3}$, the fixed initial model as reference, and sequence-summed token log-probability ratios.
Its loss averages over all prompt groups and divides each group's negative log-likelihood by its eight candidates.
Groups without a dominance comparison contribute zero loss and remain in the denominator.
This candidate-count normalization does not normalize scores by response length.
The verifier outcomes used to construct the Pareto relation are the same outcomes used for the two GRPO rewards.
Micro-batches preserve each complete eight-response group.

For MBPP, POLPO--Rate uses the same partial-order likelihood objective, policy scores, fixed initial reference, $\beta$, group-loss normalization, optimizer, learning rate, and all other hyperparameters as POLPO.
The only difference is the comparison relation: POLPO--Rate uses pass-rate rankings, so a response precedes another if and only if its mean test-pass rate is strictly higher, with equal-rate responses left unordered.

POLPO--All uses the same objective and settings as POLPO except for the learning rate and the comparison relation: a response precedes another if and only if it passes all tests and the other does not, so the responses that pass all tests are tied with one another, as are the remaining responses.
GRPO--Pareto uses the update of GRPO--All and GRPO--Rate, including its clipping and token reduction, but replaces their standardized rewards with the coefficients $\Arank{i}$ of Proposition~\ref{prop:po-gradient} evaluated at equal scores $s_1=\dots=s_N$.
These coefficients depend only on the Pareto relation and sum to zero within each group; they are not standardized, and groups without a dominance comparison receive zero coefficients.
GRPO--Pareto uses no reference model.

\subsubsection{Additional Baseline Implementations}

The following baselines are evaluated on MBPP and APPS; DPO--Pareto and GraphDPO are also evaluated on AceCode.
They use the same prompts, tests, rollout budgets, and training horizons as the corresponding GRPO and POLPO runs.

\paragraph{REINFORCE++.}
We use the original global-normalization variant of REINFORCE++ \citep{hu2025reinforcepp}, without a within-prompt mean-reward baseline.
The All and Rate variants assign the all-pass indicator and the fraction of tests passed, respectively, as terminal rewards.
With discount factor $1$, the terminal reward propagates to all response tokens; returns are then standardized across valid response tokens in the rollout batch.
The policy loss uses the same clipping parameter $0.2$ and sequence-mean token-mean reduction as GRPO, without a KL or entropy penalty.

\paragraph{GDPO.}
We adapt GDPO \citep{gdpo} to the problem-specific test sets by treating each binary test outcome as a reward component.
Each component is standardized across the eight responses to the same prompt using the sample standard deviation plus $10^{-6}$.
We average the resulting components over that problem's actual tests, broadcast the aggregate to its response tokens, and standardize across valid response tokens in the batch.
A test with a constant outcome within a response group contributes zero.
Test identities are aligned only within a prompt; absent tests are not padded into the average.
Averaging rather than summing components avoids scaling the aggregate simply by the number of tests.
The policy-loss configuration is otherwise the same as for REINFORCE++.

\paragraph{DPO--Pareto and GraphDPO.}
Both baselines use the same strict Pareto relation, fixed initial reference, sequence-summed log-probability ratios, and $\beta=10^{-3}$ as POLPO.
DPO--Pareto averages the negative log-sigmoid DPO loss over all strict dominance pairs within each prompt, including transitive pairs.
Ties and incomparable responses introduce no preference pairs.
For GraphDPO, we apply the local objective of \citet{liu2026graphdpo} to this Pareto dominance graph.
Each response that dominates at least one other response is a positive candidate in a softmax over itself and all responses it dominates, including transitive descendants.
We average these negative log-choice probabilities over such responses within the prompt.
This applies GraphDPO's objective to our verifier-derived graph; no reference solutions or ground-truth anchoring are added.
For both methods, prompt losses are averaged over all prompts, with comparison-free groups contributing zero and remaining in the denominator.
Their within-prompt pair or node averages receive no additional division by the number of candidates, whereas POLPO divides each group's negative log-likelihood by eight.

\subsubsection{Evaluation}

All coding evaluations generate one greedy response per problem and score the final checkpoint at the fixed training horizon.
The initial-model row uses the same evaluation protocol as the trained models in its table.
The response limit is 2,048 tokens for all three coding settings.
For both Qwen3 model sizes, this budget includes thinking tokens.
Training on MBPP uses a 3,072-token context, while evaluation allows a 4,096-token context to accommodate longer test prompts without reducing the 2,048-token response budget.
The APPS and AceCode evaluations use an 8,192-token context, accommodating their evaluation prompts and the 2,048-token response budget.

For MBPP and APPS, let $Q$ be the number of evaluation problems, let $K_j$ be the number of tests for problem $j$, and let $c_{jk}$ indicate whether its generated response passes test $k$.
We report
\begin{align}
  \text{All-pass}
  &=\frac{100}{Q}\sum_{j=1}^{Q}\prod_{k=1}^{K_j}c_{jk},
  \label{eq:coding-pass-one}\\
  \text{Avg. pass}
  &=\frac{100}{Q}\sum_{j=1}^{Q}\frac{1}{K_j}\sum_{k=1}^{K_j}c_{jk}.
  \label{eq:coding-mean-pass}
\end{align}
Here $Q=500$ for MBPP and $Q=1{,}000$ for APPS Introductory.
The second metric is a macro-average over problems, so problems with many tests do not dominate the score.
Compilation errors, runtime errors, and timeouts count as failures for the affected checks.

AceCode-trained models are evaluated with EvalPlus version 0.3.1, using HumanEval+ v0.1.10 (164 problems) and MBPP+ v0.2.0 (378 problems) \citep{liu2023evalplus}.
We use EvalPlus prompt formatting, code sanitization, and execution checking.
Each generated response is evaluated against both the original tests and the augmented tests; success requires passing every test in the corresponding suite.
Both columns therefore report pass@1, rather than a per-test pass rate.
The original-test MBPP score here is computed on the 378 EvalPlus problems and should not be equated with the official 500-problem MBPP test score.
Generated programs run in isolated execution environments with time limits.

\subsection{Multi-Instruction Following}
\label{app:if-details}

The IFM5 and MulDimIF experiments train separate models using the training data of the corresponding dataset and evaluate on its test data. MulDimIF results do not evaluate transfer of an IFM5-trained checkpoint. For IFM5 and MulDimIF, the initial-model row denotes Qwen3-4B-Instruct-2507 before the additional training in these experiments.\footnote{\url{https://huggingface.co/Qwen/Qwen3-4B-Instruct-2507}}

\subsubsection{IFM5 Data and Split}
\label{app:ifm5-data}

We use \texttt{allenai/IF\_multi\_constraints\_upto5}, released with \citet{pyatkin2025generalizing}.\footnote{\url{https://huggingface.co/datasets/allenai/IF_multi_constraints_upto5}} The dataset card identifies this as IF-RLVR training data with up to five constraints per prompt, drawn from 25 IFEval and 29 IFBench-Train constraint templates. We abbreviate this dataset as IFM5. Our test set is a holdout from this released training dataset, not the original IFBench benchmark test set.

All 95,373 rows are uniformly shuffled, the first 2,000 rows are assigned to the test split, and the remaining 93,373 rows to the training split. This is a row-level random split without stratification by the number of constraints. Table~\ref{tab:ifm5-split} gives the test distribution before and after length filtering. The initial distribution is close to the reported approximate training proportions of 24\%, 25\%, 24\%, 19\%, and 7\% for $K=1,\ldots,5$, respectively; these training percentages are rounded summaries, not exact counts.

\begin{table}[ht]
\centering
\caption{IFM5 test counts before and after excluding prompts longer than 2,048 tokens. The retained counts are the denominators for the reported complete satisfaction rates and the weights for overall aggregation.}
\label{tab:ifm5-split}
\begin{tabular}{lrrrrrr}
\toprule
Constraint count $K$ & 1 & 2 & 3 & 4 & 5 & Total \\
\midrule
Before filtering & 478 & 493 & 500 & 394 & 135 & 2,000 \\
Excluded & 2 & 5 & 7 & 3 & 0 & 17 \\
Retained & 476 & 488 & 493 & 391 & 135 & 1,983 \\
\bottomrule
\end{tabular}
\end{table}

Prompts exceeding 2,048 tokens are excluded as entire examples, not shortened. This excludes 17 test prompts, leaving 1,983 for evaluation.

\subsubsection{MulDimIF Data and Levels}
\label{app:muldimif-data}

Models in the MulDimIF experiment are trained on its training split and evaluated on its test split \citep{ye2025muldimif}. The paper defines four levels using both the number of constraint categories and the number of individual constraints (Table~\ref{tab:muldimif-levels}). Thus, Level IV does not mean exactly four individual constraints. The framework also distinguishes example, listing, and incorporation presentation patterns.

\begin{table}[ht]
\centering
\caption{MulDimIF difficulty levels, following \citet[Section~3.1.3]{ye2025muldimif}.}
\label{tab:muldimif-levels}
\begin{tabular}{lcc}
\toprule
Level & Constraint categories & Individual constraints \\
\midrule
I & 1 & 1--2 \\
II & 2 & 2--4 \\
III & 3 & 3--6 \\
IV & 4 & 4--8 \\
\bottomrule
\end{tabular}
\end{table}

\subsubsection{Verification and Evaluation Metrics}
\label{app:if-metrics}

For evaluated prompt $x_j$, let $K_j$ be its number of constraints and $c_{jk}\in\{0,1\}$ indicate whether the generated response passes check $k$. Complete constraint satisfaction is
\begin{equation}
u_j=\prod_{k=1}^{K_j}c_{jk},
\qquad
\mathrm{CS}=\frac{1}{n}\sum_{j=1}^{n}u_j,
\label{eq:complete-satisfaction}
\end{equation}
where $n$ is the number of evaluated prompts after applicable exclusions. Tables report $100\,\mathrm{CS}$ as a percentage.

For IFM5, let $n_k$ count retained prompts with $K_j=k$ and let $A_k$ be their mean complete satisfaction. The overall rate is the prompt-count-weighted average
\begin{equation}
\mathrm{CS}_{\mathrm{IFM5}}
=\frac{\sum_{k=1}^{5}n_k A_k}{\sum_{k=1}^{5}n_k}.
\label{eq:ifm5-overall}
\end{equation}
The weights are the retained counts in Table~\ref{tab:ifm5-split}: $(n_1,\ldots,n_5)=(476,488,493,391,135)$. Overall values are retained from the evaluation summaries rather than recomputed from rounded subgroup rates.

For both IFM5 and MulDimIF, each trained method is reported as the mean and standard deviation across three training seeds. IFM5 test responses use greedy decoding.

\subsubsection{Training Configurations}
\label{app:if-training}

Table~\ref{tab:if-hyperparameters} summarizes the training hyperparameters for IFM5 and MulDimIF.
Both configurations use 64 prompts per rollout batch and 16 sampled responses per prompt, giving 1,024 responses per full rollout batch.
The configured training horizon is 400 trainer steps for each dataset.

\begin{table}[t]
\centering
\small
\caption{Training hyperparameters for IFM5 and MulDimIF. Rollout batch size counts prompts. The training horizon is measured in trainer steps; the update mini-batch size is reported as configured. Sampling settings apply to training rollouts.}
\label{tab:if-hyperparameters}
\begin{tabular}{lcc}
\toprule
Setting & IFM5 & MulDimIF \\
\midrule
Initial model & \multicolumn{2}{c}{Qwen3-4B-Instruct-2507} \\
Optimizer & AdamW & AdamW \\
Learning rate & $10^{-6}$ & $10^{-6}$ \\
Adam coefficients $(\beta_1,\beta_2)$ & $(0.9,0.999)$ & $(0.9,0.999)$ \\
Weight decay & $0.01$ & $0.01$ \\
Learning-rate schedule & Constant & Constant \\
Warmup & None & None \\
Gradient clipping norm & $1.0$ & $1.0$ \\
Prompts per rollout batch & 64 & 64 \\
Responses per prompt & 16 & 16 \\
Update mini-batch size & 16 & 16 \\
Update epochs per rollout batch & 1 & 1 \\
Training horizon (trainer steps) & 400 & 400 \\
Maximum prompt length (tokens) & 2,048 & 4,096 \\
Maximum response length (tokens) & 2,048 & 4,096 \\
Sampling temperature & $1.0$ & $1.0$ \\
Top-$p$ & $1.0$ & $1.0$ \\
Top-$k$ & Disabled & Disabled \\
\bottomrule
\end{tabular}
\end{table}

On both IFM5 and MulDimIF, we compare the initial model, GRPO--All, GRPO--Rate, GDPO \citep{gdpo}, GraphDPO, DPO--Pareto, and POLPO (Tables~\ref{tab:ifm5-overall} and~\ref{tab:muldimif-overall}). GraphDPO and DPO--Pareto use the Pareto relation, and the reported GraphDPO, DPO--Pareto, and POLPO results use the non-normalized variants.
GRPO--All assigns reward one only when every constraint is satisfied, whereas GRPO--Rate uses the fraction of satisfied constraints, as defined in \eqn{eq:reward-baselines}.
Both GRPO--All and GRPO--Rate use a KL coefficient of $10^{-3}$.

POLPO constructs comparisons from the individual pass/fail vectors using \eqn{eq:pareto} and optimizes the partial-order likelihood in \eqn{eq:polpo} with score scale $\beta=10^{-3}$.
The GRPO KL coefficient controls the KL penalty, whereas POLPO's $\beta$ scales the policy-to-reference scores; their equal numerical values do not imply equal regularization strength.

\subsubsection{MathIF: Data, Training, and Evaluation}
\label{app:mathif-details}

\paragraph{Benchmark and task.}
MathIF combines mathematical questions with one, two, or three programmatically verifiable instruction constraints \citep{fu2025scaling}.
We evaluate all 420 examples: GSM8K, MATH-500, Minerva, and OlympiadBench each contribute 30 examples at each constraint count, while AIME 2024/25 contributes 20, giving 140 examples at each count.
The benchmark uses 15 constraint types covering lexical, length, format, language, and affix requirements.
The prompt contains the mathematical question, a request for step-by-step reasoning and a final answer in \verb|\boxed{}|, and the constraint descriptions, in a single user turn without a system prompt.

\paragraph{Training data and overlap filtering.}
We construct a separate training dataset by adding constraints to problems from the MATH training split \citep{hendrycks2021math}, using \nolinkurl{DigitalLearningGmbH/MATH-lighteval} as the source release.
Reference answers are extracted from the final boxed expression in each supplied solution.
We filter source problems against both the 420 MathIF examples and MATH-500 using normalized exact matching, or a shared word 10-gram together with a \texttt{difflib} similarity ratio of at least $0.9$.
Problems sharing a template but differing only in numerical values are retained; this filtering therefore does not establish the absence of all semantic overlap.
Of the 7,500 source problems, 55 are removed by overlap filtering, two because their question text exceeds 3,000 characters, and two because their solutions contain no boxed answer that can be extracted.
These exclusions leave 7,441 source problems, split into 6,941 for training and 500 for validation.
Each problem yields one example at each of $K=1,2,3$ constraints, producing 20,823 training examples and 1,500 validation examples.
An additional validation set contains the 500 MATH-500 problems without added constraints.
No difficulty-based or all-pass filtering is applied.

\paragraph{Constraint generation.}
We use the constraint wording and sample constraint parameters from the MathIF implementation included in our codebase.
Training uses 14 of the benchmark's 15 constraint types, excluding the requirement to answer in a specified language.
Each training prompt follows the same format as an evaluation prompt: the mathematical question, a request for step-by-step reasoning, and the sampled constraints.

We check which constraints can be combined before sampling them together.
If MathIF lists two constraint types as incompatible in either order, we treat the pair as incompatible in both orders.
For the constraint that requires repeating the prompt, we adjust this list to permit the relevant benchmark combinations: its incompatible types are quotation, all-uppercase, all-lowercase, multiple-section, and response-language requirements.
We also exclude combinations of all-lowercase text with multiple sections, and combinations of multiple sections with a bullet list.
When all-uppercase text and multiple sections are required together, we use \texttt{SECTION} as the section delimiter.

We also check the question and reference answer when their contents could conflict with a constraint.
A no-comma constraint is sampled only if the reference answer contains no comma; if the prompt must be repeated, the question must also contain no comma.
Before sampling combinations that require repeating the prompt, we check the question for forbidden words and ensure that repeating it would not already exceed the specified word-frequency or word-count limits.
Keywords are distinct within each set of constraints.

\paragraph{Verification and training objectives.}
For a response to a problem with $K$ instruction constraints, the feedback vector is
\begin{equation}
\mathbf{v}=(a,c_1,\ldots,c_K)\in\{0,1\}^{K+1},
\qquad
r_{\mathrm{rate}}=\frac{a+\sum_{k=1}^{K}c_k}{K+1},
\label{eq:mathif-feedback}
\end{equation}
where $a$ indicates mathematical correctness and $c_k$ indicates satisfaction of constraint $k$.
GRPO--All uses $r_{\mathrm{all}}=a\prod_{k=1}^{K}c_k$, which is one only when the response is mathematically correct and satisfies every instruction constraint.
GRPO--Rate uses $r_{\mathrm{rate}}$ as its scalar reward, while POLPO constructs the dominance relation from the full vector, including correctness. The thinking-enabled GraphDPO and DPO--Pareto results use the same Pareto rule and non-normalized variants; the reported POLPO result also uses the non-normalized variant.
We use the same verifier implementations during training and evaluation for both mathematical correctness and instruction-constraint satisfaction.
Mathematical correctness is evaluated with the mathematical-answer grader distributed with \texttt{verl} and used by MathIF.
In the thinking-enabled runs, a response whose thinking segment is not closed receives zero for the mathematical correctness bit and every constraint bit; this rule applies to both training feedback and evaluation.
The thinking-enabled results use our verification rules, including LaTeX-aware case checks for instruction constraints, rather than the original MathIF text rules.

\paragraph{Optimization.}
Table~\ref{tab:mathif-hyperparameters} summarizes the MathIF training hyperparameters in the same format as Table~\ref{tab:if-hyperparameters}.
The original non-thinking MathIF training runs initialize \nolinkurl{Qwen/Qwen3-0.6B} with thinking disabled.
In these non-thinking runs, each rollout batch contains 64 prompts with eight responses per prompt, giving 512 sampled responses.
The evaluated trained-model checkpoints correspond to the reported one-epoch runs of 325 trainer steps.
Both GRPO--All and GRPO--Rate use a KL coefficient of $10^{-3}$; POLPO uses score scale $\beta=10^{-3}$ and no additional KL penalty.

\begin{table}[t]
\centering
\small
\caption{MathIF training hyperparameters by thinking mode. Rollout and update mini-batch sizes count prompts; sampling settings apply to training rollouts. Shared values apply to both modes.}
\label{tab:mathif-hyperparameters}
\begin{tabular}{lcc}
\toprule
Setting & Thinking disabled & Thinking enabled \\
\midrule
Initial model & \multicolumn{2}{c}{Qwen3-0.6B} \\
Optimizer & \multicolumn{2}{c}{AdamW} \\
Learning rate & \multicolumn{2}{c}{$10^{-6}$} \\
Adam coefficients $(\beta_1,\beta_2)$ & \multicolumn{2}{c}{$(0.9,0.999)$} \\
Weight decay & \multicolumn{2}{c}{$0.01$} \\
Learning-rate schedule & \multicolumn{2}{c}{Constant} \\
Warmup & \multicolumn{2}{c}{None} \\
Gradient clipping norm & \multicolumn{2}{c}{$1.0$} \\
Prompts per rollout batch & \multicolumn{2}{c}{64} \\
Responses per prompt & \multicolumn{2}{c}{8} \\
Update mini-batch size & \multicolumn{2}{c}{16} \\
Update epochs per rollout batch & \multicolumn{2}{c}{1} \\
Training epochs & \multicolumn{2}{c}{1} \\
Maximum prompt length (tokens) & \multicolumn{2}{c}{2,048} \\
Maximum response length (tokens) & 2,048 & 8,192 \\
Sampling temperature & \multicolumn{2}{c}{$1.0$} \\
Top-$p$ & \multicolumn{2}{c}{$1.0$} \\
Top-$k$ & \multicolumn{2}{c}{Disabled} \\
\bottomrule
\end{tabular}
\end{table}

\paragraph{Evaluation and metrics.}
In both thinking modes, evaluation uses one greedy response per MathIF example with a maximum generation length of 16,384 tokens. In thinking-enabled evaluation, this budget includes the thinking segment and the final answer; the training rollout response limit remains 8,192 tokens (Table~\ref{tab:mathif-hyperparameters}).
Benchmark source and constraint count are identified from the dataset filenames because some record identifiers do not match those attributes.
For the $Q=420$ evaluated examples, let $a_j$ be mathematical correctness, $K_j$ the number of instruction constraints, and $c_{jk}$ the evaluation checker's outcome for constraint $k$.
We report
\begin{align}
\mathrm{HAcc}&=\frac{100}{Q}\sum_{j=1}^{Q}\prod_{k=1}^{K_j}c_{jk},
&\mathrm{SAcc}&=\frac{100}{Q}\sum_{j=1}^{Q}\frac{1}{K_j}\sum_{k=1}^{K_j}c_{jk},\\
\mathrm{Acc}&=\frac{100}{Q}\sum_{j=1}^{Q}a_j,
&\mathrm{Joint}&=\frac{100}{Q}\sum_{j=1}^{Q}a_j\prod_{k=1}^{K_j}c_{jk}.
\label{eq:mathif-metrics}
\end{align}
HAcc and SAcc measure instruction adherence without requiring mathematical correctness; Joint measures their conjunction with correctness.
All displayed percentages retain the precision of the evaluation summaries.

\section{Additional Experiments}
\label{app:additional-experiments}

\subsection{Measured Toy Learning Efficiency}
\label{app:toy-efficiency}

\begin{figure}[t]
\centering
\includegraphics[width=\linewidth]{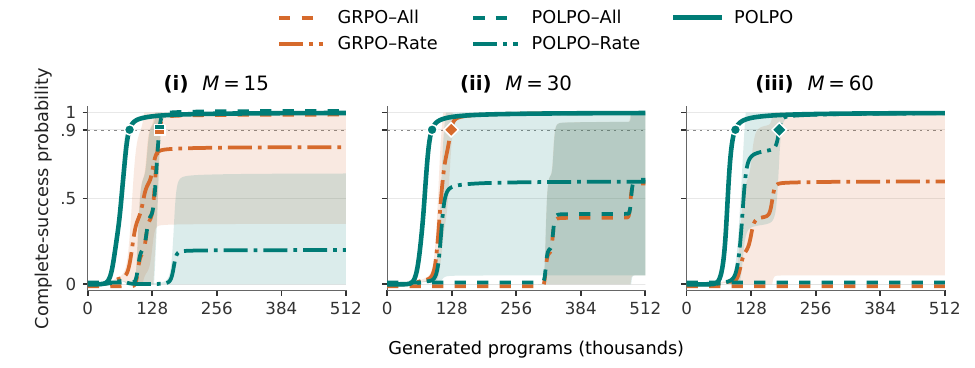}
\caption{Complete-success probability for target $(3,5)$ with search spaces $M=15$, $30$, and $60$ over the full budget of 512,000 programs, averaged over five training seeds.
Shading shows one sample standard deviation across seeds, clipped to $[0,1]$.
The horizontal dotted line marks $0.9$, and markers indicate the first recorded crossings of the mean curves.}
\label{fig:toy-maximum-sweep}
\end{figure}

Table~\ref{tab:toy-efficiency} summarizes the threshold crossings in Figures~\ref{fig:toy-learning} and~\ref{fig:toy-maximum-sweep}.
The values are computed separately from each of the five saved training trajectories.
We report the mean and sample standard deviation of the first recorded crossing budgets only when all five runs reach the threshold.
Otherwise, the table gives the number of runs that reach it, so that runs without a crossing are not silently excluded.

Figure~\ref{fig:toy-maximum-sweep} shows the first sweep, which varies the search space for target $(3,5)$ over the full budget.
POLPO reaches the threshold in all five seeds for every $M$, using on average 82.0, 88.7, and 96.5 thousand programs for $M=15$, $30$, and $60$, respectively.
GRPO--All and POLPO--All reach it in all five seeds for $M=15$, in three seeds for $M=30$, and in no seed for $M=60$, and their curves nearly coincide.
POLPO--Rate reaches it in one, three, and five seeds, and GRPO--Rate in four, five, and three seeds, respectively.

\begin{table}[t]
\centering
\small
\caption{Generated programs needed to reach $0.9$ complete-success probability, in thousands (mean $\pm$ sample standard deviation over five seeds).
The first two rows vary the search space $M$ for target $(3,5)$; the last three use $M=60$, as in Figure~\ref{fig:toy-learning}.
A dash with $(k/5)$ indicates that only $k$ runs reached the threshold within 512,000 programs, so a mean crossing budget is not reported.
Crossings are taken from every-update records without interpolation.}
\label{tab:toy-efficiency}
\setlength{\tabcolsep}{4pt}
\begin{tabular}{lcccccc}
\toprule
Target & $M$ & GRPO--All & GRPO--Rate & POLPO--All & POLPO--Rate & POLPO \\
\midrule
$(3,5)$  & 15 & $131.0\pm16.2$ & --- $(4/5)$ & $132.0\pm16.3$ & --- $(1/5)$ & $\mathbf{82.0\pm3.3}$ \\
$(3,5)$  & 30 & --- $(3/5)$ & $118.7\pm8.8$ & --- $(3/5)$ & --- $(3/5)$ & $\mathbf{88.7\pm2.2}$ \\
\midrule
$(3,5)$  & 60 & --- $(0/5)$ & --- $(3/5)$ & --- $(0/5)$ & $136.5\pm30.3$ & $\mathbf{96.5\pm0.6}$ \\
$(3,7)$  & 60 & --- $(0/5)$ & --- $(3/5)$ & --- $(0/5)$ & $151.0\pm13.8$ & $\mathbf{99.5\pm4.6}$ \\
$(3,11)$ & 60 & --- $(0/5)$ & --- $(1/5)$ & --- $(0/5)$ & --- $(3/5)$ & $\mathbf{99.7\pm1.9}$ \\
\bottomrule
\end{tabular}
\end{table}

\begin{figure}[t]
\centering
\includegraphics[width=\linewidth]{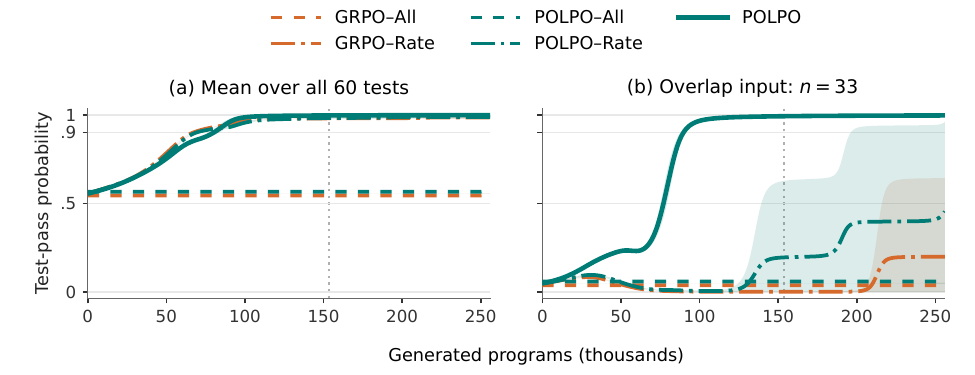}
\caption{Diagnostic curves for target $(3,11)$, averaged over five training seeds, with shading showing one sample standard deviation clipped to $[0,1]$.
The left panel averages pass probability over all 60 verifiers, and the right panel shows the overlap input $n=33$.
The vertical dotted line marks 153,600 generated programs.}
\label{fig:toy-overlap-appendix}
\end{figure}

\subsection{Learning-Signal Availability in Coding Tasks}
\label{app:mbpp-signal}

POLPO compares two responses only when one Pareto-dominates the other, so on the same responses it obtains a learning signal from at most as many groups as a pass-rate reward.
This is intended: a pass-rate reward also orders responses with conflicting test outcomes, which prescribes a trade-off between tests that verification does not establish and can steer learning away from complete solutions (Section~\ref{sec:exp-toy}).
A smaller share of groups with a learning signal is therefore not a drawback in itself.
Because a group provides no signal under any of the compared conditions once all eight responses pass every test, we report both the share of training groups that provide a learning signal and the same share among unsolved groups, in which not all eight responses pass every test.

\begin{figure}[t]
\centering
\includegraphics[width=\linewidth]{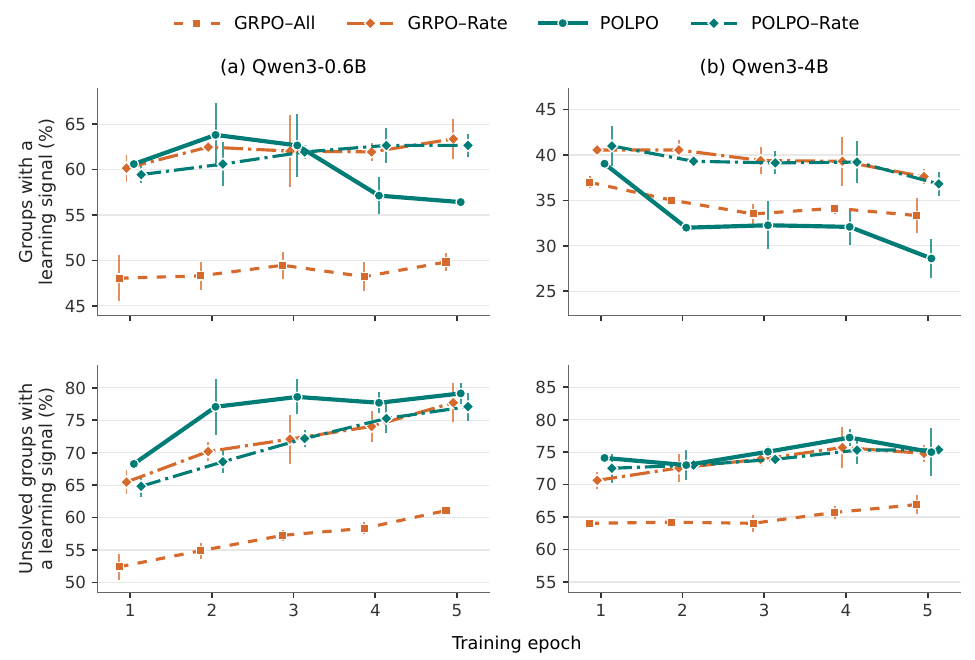}
\caption{Learning signal on MBPP in each training epoch.
Top: share of training groups that provide a learning signal.
A group is the eight responses sampled for one problem at one update; it provides a signal when it contains a strict Pareto-dominance pair (POLPO), at least two distinct test-pass rates (POLPO--Rate and GRPO--Rate), or both a response that passes every test and one that does not (GRPO--All), and otherwise contributes no policy gradient.
Bottom: the same share among unsolved groups, in which not all eight responses pass every test.
Points show the mean over three training seeds of the runs in Table~\ref{tab:mbpp-coding}, and bars show one sample standard deviation.
Within each row, the two panels use the same vertical scale over different ranges.}
\label{fig:mbpp-signal}
\end{figure}

Figure~\ref{fig:mbpp-signal} shows both shares for MBPP in each training epoch.
For Qwen3-0.6B, POLPO, POLPO--Rate, and GRPO--Rate obtain a signal from roughly 55--65\% of groups, whereas GRPO--All stays below 50\%.
For Qwen3-4B, the shares are lower for all methods, ranging from about 29\% to 41\%.
In the first epoch, when all runs start from the same initial model, POLPO's share is within two percentage points of those of POLPO--Rate and GRPO--Rate and above that of GRPO--All for both models.
Later in training, POLPO's share decreases, to about 56--57\% in the last two epochs for Qwen3-0.6B and to about 29--32\% from the second epoch for Qwen3-4B, but it stays above that of GRPO--All for Qwen3-0.6B and within five percentage points of it for Qwen3-4B.
Among unsolved groups, POLPO's share shows no such decrease: for Qwen3-4B, it stays between about 73\% and 77\%, close to those of POLPO--Rate and GRPO--Rate, and for Qwen3-0.6B, it is the highest of the four methods in every epoch.

\begin{figure}[t]
\centering
\includegraphics[width=\linewidth]{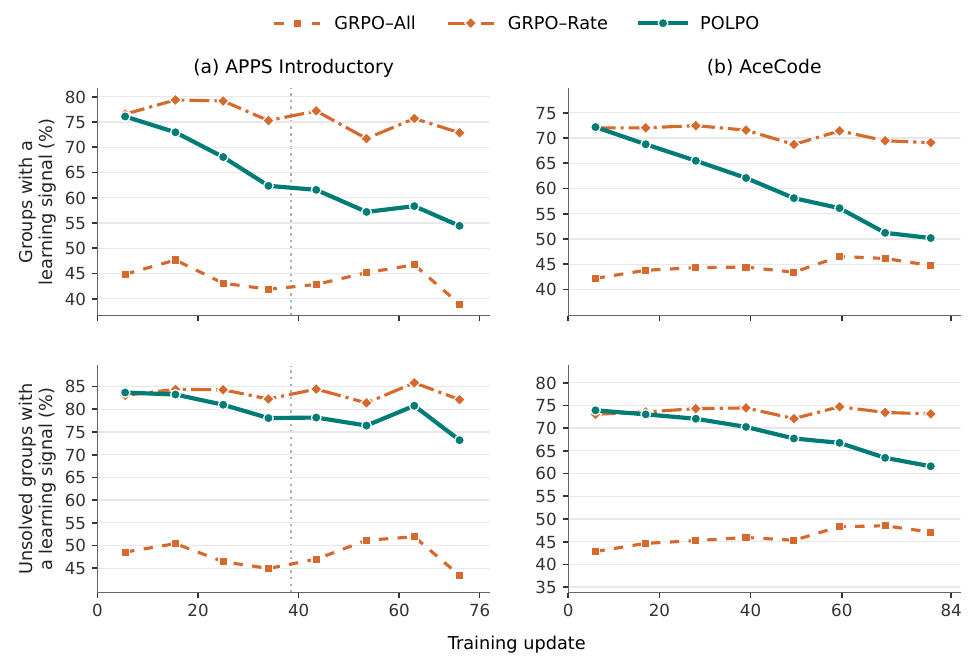}
\caption{Learning signal on APPS Introductory and AceCode during training.
Top: share of training groups that provide a learning signal; bottom: the same share among unsolved groups, as in Figure~\ref{fig:mbpp-signal}.
Each point pools about ten consecutive updates of the training run in Table~\ref{tab:apps-coding} or Table~\ref{tab:acecode-coding}, and the dotted line marks the boundary between the two APPS Introductory epochs.
Within each row, the two panels use the same vertical scale over different ranges.}
\label{fig:code-signal}
\end{figure}

Figure~\ref{fig:code-signal} shows the same shares for APPS Introductory and AceCode, for the training runs in Tables~\ref{tab:apps-coding} and~\ref{tab:acecode-coding}.
POLPO starts at the level of GRPO--Rate, with about 76\% of groups for APPS Introductory and 72\% for AceCode, and decreases over training to about 54\% and 50\%, respectively, while GRPO--Rate stays between about 69\% and 79\% and GRPO--All between about 39\% and 48\%.
Among unsolved groups, POLPO's share decreases less, from about 84\% to 73\% for APPS Introductory and from about 74\% to 62\% for AceCode; it remains well above that of GRPO--All, which is at most about 52\%, and falls below that of GRPO--Rate later in training.
Thus, restricting comparisons to verified improvements still leaves POLPO with a learning signal from more unsolved groups than GRPO--All in all three datasets, and from a share comparable to that of GRPO--Rate on MBPP and early in training on APPS Introductory and AceCode.

\subsection{Structure of the Pareto Partial Orders}
\label{app:partial-order-structure}

\begin{figure}[t]
\centering
\includegraphics[width=\linewidth]{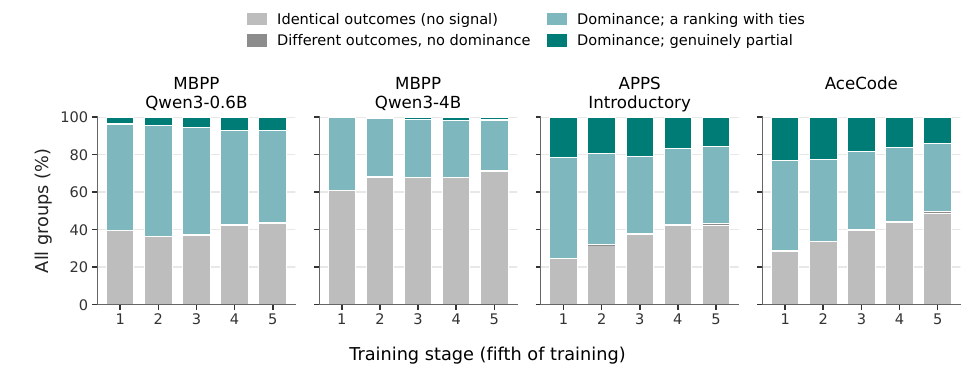}
\caption{Structure of the Pareto partial orders in POLPO's training groups for the coding tasks, per fifth of training (epochs for MBPP, fifths of the updates for APPS Introductory and AceCode).
All groups are divided by whether the eight responses have identical test outcomes, differ without any dominance pair, are ordered as a ranking with ties, or form a genuinely partial order containing two incomparable responses with different outcomes.
MBPP pools the three training seeds of the runs in Table~\ref{tab:mbpp-coding}; APPS Introductory and AceCode show the runs in Tables~\ref{tab:apps-coding} and~\ref{tab:acecode-coding}.}
\label{fig:partial-order-structure}
\end{figure}

Figure~\ref{fig:partial-order-structure} characterizes the partial orders that POLPO receives in the coding tasks, using the same training groups as Figures~\ref{fig:mbpp-signal} and~\ref{fig:code-signal}.
It divides all groups into four kinds: the eight responses have identical test outcomes; their outcomes differ but contain no dominance pair; their Pareto order is a ranking with ties; or their Pareto order is genuinely partial, containing two responses with different outcomes neither of which dominates the other.
Groups whose outcomes differ without any dominance pair are rare, at most about 1\% of groups in every dataset and stage.
Genuinely partial orders occur in about 14--23\% of all groups for APPS Introductory and AceCode, compared with at most about 7\% for Qwen3-0.6B and 2\% for Qwen3-4B on MBPP, where each problem has three tests.
Over training, the share of groups with identical outcomes increases in every dataset.

\begin{figure}[t]
\centering
\includegraphics[width=\linewidth]{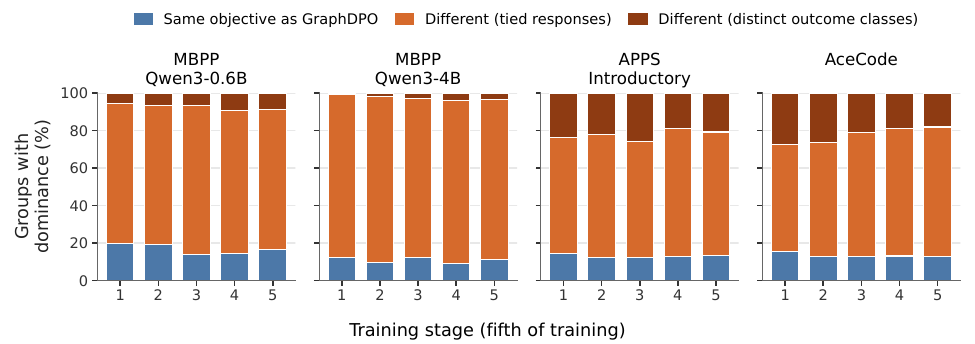}
\caption{Comparison of POLPO's objective with GraphDPO's on the training groups of Figure~\ref{fig:partial-order-structure} that contain dominance, per fifth of training.
The groups are divided by whether POLPO's objective equals GraphDPO's up to a constant factor (every response has at most one immediate dominator) or differs because tied responses or responses with distinct outcomes dominate the same response.}
\label{fig:partial-order-graphdpo}
\end{figure}

Figure~\ref{fig:partial-order-graphdpo} compares POLPO's objective with GraphDPO's on the groups with dominance.
When every response has at most one immediate dominator, so that the dominance relation forms a rooted forest, the Plackett--Luce events that each dominating response precedes its descendants are independent.
POLPO's group log-likelihood then equals the sum of GraphDPO's per-response log-softmax terms, and the two objectives differ only by a constant factor.
Otherwise, they are different functions of the scores, which is the case for about 80--90\% of the groups with dominance in every dataset.
Most of these groups contain several responses with the same outcomes that dominate the same responses: POLPO requires each dominated response to follow all of them jointly, whereas GraphDPO adds a separate softmax term for each dominating response.
For APPS Introductory and AceCode, about 18--27\% of the groups with dominance also contain responses with distinct outcomes that dominate the same response.

\subsection{Learning Signals and Pareto-Order Structure in Instruction Following}
\label{app:if-rollout-analysis}

\begin{figure}[t]
\centering
\includegraphics[width=\linewidth]{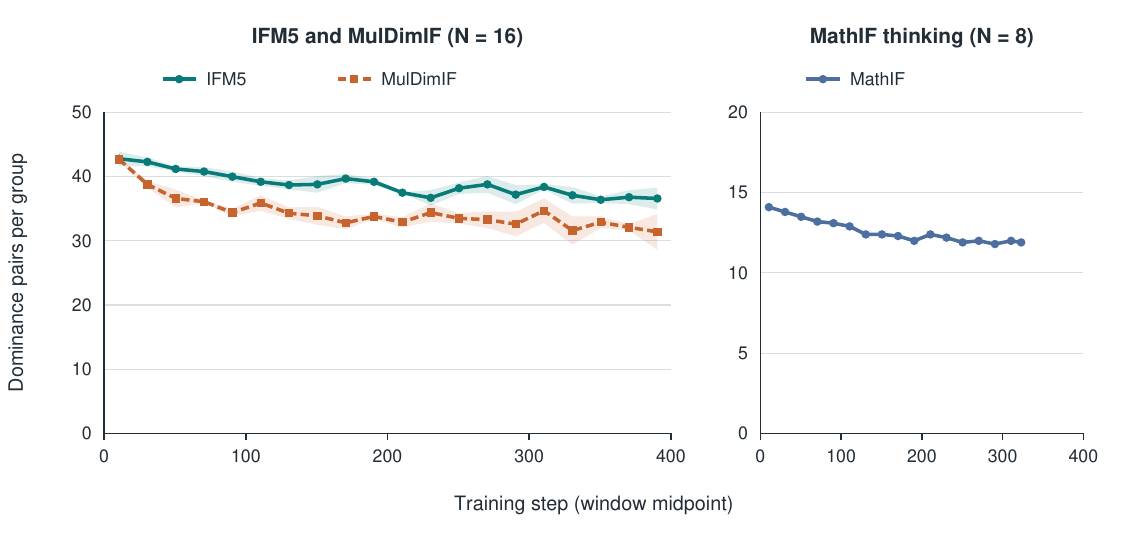}
\caption{Mean number of strict dominance pairs per group containing at least one dominance pair, in instruction-following training rollouts. Each point represents a 20-step window; the last MathIF point covers only steps 321--325 and is placed at their midpoint. IFM5 and MulDimIF results show the mean over three training seeds, with shading indicating $\pm$ one standard deviation across seeds; MathIF uses one seed. The panels have different vertical scales. The maximum possible pair count is 120 for $N=16$ and 28 for $N=8$.}
\label{fig:if-dominance-pairs}
\end{figure}

We extend the rollout analyses of Appendices~\ref{app:mbpp-signal} and~\ref{app:partial-order-structure} to IFM5, MulDimIF, and MathIF with thinking enabled. We analyze initial-policy and final-policy response groups: 988 groups for IFM5, 800 for MulDimIF, and 375 for MathIF. A group contains responses to a prompt. The final groups for each method come from the policy trained by that method; thus, final-method comparisons reflect both the objective and the policy's sampled responses. These endpoint summaries differ from the single-response benchmark scores in Section~\ref{sec:exp-multi}. We also report the number of dominance pairs in training rollouts over time in Figure~\ref{fig:if-dominance-pairs}.

\paragraph{Learning-signal availability.}
As in the coding analysis, a group supplies a POLPO signal if it contains a strict Pareto-dominance pair, a GRPO--Rate signal if its responses have different fractions of satisfied checks, and a GRPO--All signal if it contains both a fully successful response and a response that is not fully successful. For MathIF, the checks include mathematical correctness as well as the instruction constraints (\eqn{eq:mathif-feedback}). A group is \emph{solved} here only when every response in it satisfies every check; an unsolved group is any other group. This group-level notion differs from the single-response success metrics in the main experiments.

Table~\ref{tab:if-rollout-signal} reports signal shares over all groups and over unsolved groups. At initialization, POLPO obtains a signal from 47.7\%, 74.8\%, and 75.2\% of groups on IFM5, MulDimIF, and MathIF, respectively, compared with 17.7\%, 43.0\%, and 26.4\% for GRPO--All and 47.7\%, 74.8\%, and 75.5\% for GRPO--Rate. At the POLPO endpoint, the all-group signal shares decrease to 14.0\%, 11.1\%, and 52.5\%, while the solved shares increase to 56.5\%, 84.4\%, and 38.9\%. Among final unsolved groups, POLPO obtains a signal from 32.1\%, 71.2\%, and 86.0\% of groups, compared with 27.2\%, 64.3\%, and 67.4\% for GRPO--All and 41.5\%, 80.1\%, and 86.1\% for GRPO--Rate. These final comparisons use each method's own rollouts and therefore reflect both differences in signal criteria and differences in sampled responses. The all-group and unsolved-group views provide complementary information.

\begin{table}[htbp]
\centering
\small
\setlength{\tabcolsep}{4pt}
\caption{Learning-signal availability (\%) in instruction-following rollout groups. Each method uses its own signal criterion. Initial and final signal shares use all groups or only unsolved groups; solved shares are over all groups. Final groups are sampled from the corresponding method's trained policy.}
\label{tab:if-rollout-signal}
\begin{tabular}{@{}lrrrrr@{}}
\toprule
& \multicolumn{2}{c}{Initial signal} & \multicolumn{2}{c}{Final signal} & Solved \\
\cmidrule(lr){2-3}\cmidrule(lr){4-5}
Method & All & Unsolved & All & Unsolved & Initial $\to$ final \\
\midrule
\multicolumn{6}{@{}l}{\emph{IFM5} ($n=988$)} \\
POLPO     & 47.7 & 57.7 & 14.0 & 32.1 & $17.4\to56.5$ \\
GRPO--All  & 17.7 & 21.4 & 13.1 & 27.2 & $17.4\to51.9$ \\
GRPO--Rate & 47.7 & 57.7 & 19.1 & 41.5 & $17.4\to53.9$ \\
\midrule
\multicolumn{6}{@{}l}{\emph{MulDimIF} ($n=800$)} \\
POLPO     & 74.8 & 89.0 & 11.1 & 71.2 & $16.0\to84.4$ \\
GRPO--All  & 43.0 & 51.2 & 12.4 & 64.3 & $16.0\to80.8$ \\
GRPO--Rate & 74.8 & 89.0 & 15.1 & 80.1 & $16.0\to81.1$ \\
\midrule
\multicolumn{6}{@{}l}{\emph{MathIF, thinking enabled} ($n=375$)} \\
POLPO     & 75.2 & 79.2 & 52.5 & 86.0 & $5.1\to38.9$ \\
GRPO--All  & 26.4 & 27.8 & 43.5 & 67.4 & $5.1\to35.5$ \\
GRPO--Rate & 75.5 & 79.5 & 59.5 & 86.1 & $5.1\to30.9$ \\
\bottomrule
\end{tabular}
\end{table}

\paragraph{Pareto-order structure.}
Table~\ref{tab:if-rollout-structure} partitions the initial and POLPO-final groups using the same categories as Figure~\ref{fig:partial-order-structure}. The share of genuinely partial orders is 8.9\% on initial IFM5, 31.2\% on initial MulDimIF, and 29.3\% on initial MathIF; it falls to 1.4\%, 0.5\%, and 11.5\%, respectively, at the POLPO endpoint. Meanwhile, the share of groups with identical outcome vectors rises from 52.1\% to 84.7\% on IFM5, from 25.1\% to 88.9\% on MulDimIF, and from 23.5\% to 47.2\% on MathIF. Identical outcomes alone do not imply that all responses are fully successful. Groups with differing outcomes but no dominance pair account for at most 1.3\% of groups in these snapshots.

\begin{table}[htbp]
\centering
\small
\setlength{\tabcolsep}{4pt}
\caption{Structure of initial-policy and POLPO-final instruction-following rollout groups (\% of all groups). Identical means that all verification vectors match. The second category has different vectors but no dominance pair. The remaining categories contain dominance: the distinct vectors form a ranking with ties, or the order is genuinely partial, with incomparable distinct vectors. Percentages may not sum to 100 because of rounding.}
\label{tab:if-rollout-structure}
\begin{tabular}{@{}lrrrrr@{}}
\toprule
Policy & $n$ & \shortstack{Identical\\outcomes} & \shortstack{Different,\\no dominance} & \shortstack{Ranking\\with ties} & \shortstack{Genuinely\\partial} \\
\midrule
\multicolumn{6}{@{}l}{\emph{IFM5}} \\
Initial     & 988 & 52.1 & 0.2 & 38.8 & 8.9 \\
POLPO final & 988 & 84.7 & 1.3 & 12.6 & 1.4 \\
\midrule
\multicolumn{6}{@{}l}{\emph{MulDimIF}} \\
Initial     & 800 & 25.1 & 0.1 & 43.5 & 31.2 \\
POLPO final & 800 & 88.9 & 0.0 & 10.6 & 0.5 \\
\midrule
\multicolumn{6}{@{}l}{\emph{MathIF, thinking enabled}} \\
Initial     & 375 & 23.5 & 1.3 & 45.9 & 29.3 \\
POLPO final & 375 & 47.2 & 0.3 & 41.1 & 11.5 \\
\bottomrule
\end{tabular}
\end{table}

\paragraph{POLPO and GraphDPO objectives.}
Table~\ref{tab:if-rollout-graphdpo} compares the objectives on the subset of groups that contain dominance, following Figure~\ref{fig:partial-order-graphdpo}. Across the six dataset--policy snapshots, the objectives agree up to a constant factor on 2.2--11.0\% of dominance groups; they differ on the remainder. Differences attributed to tied responses account for 61.7--95.5\% of dominance groups. Differences attributed to distinct outcome classes are more frequent initially (12.3\%, 22.9\%, and 27.3\% on IFM5, MulDimIF, and MathIF) than at the POLPO endpoint (5.8\%, 1.1\%, and 7.6\%). These shares describe objective structure on sampled groups, rather than measured differences in model performance.

\begin{table}[htbp]
\centering
\small
\setlength{\tabcolsep}{5pt}
\caption{POLPO versus GraphDPO objectives on instruction-following groups with at least one dominance pair. Percentages use the number of dominance groups in each row as the denominator. ``Same'' means equal up to a constant factor; the other columns distinguish differences attributed to tied responses and to distinct outcome classes, as in Figure~\ref{fig:partial-order-graphdpo}.}
\label{tab:if-rollout-graphdpo}
\begin{tabular}{@{}lrrrr@{}}
\toprule
Policy & \shortstack{Dominance\\groups} & Same & \shortstack{Different:\\tied responses} & \shortstack{Different:\\distinct outcome classes} \\
\midrule
\multicolumn{5}{@{}l}{\emph{IFM5}} \\
Initial     & 471 & 7.2 & 80.5 & 12.3 \\
POLPO final & 138 & 5.8 & 88.4 & 5.8 \\
\midrule
\multicolumn{5}{@{}l}{\emph{MulDimIF}} \\
Initial     & 598 & 2.2 & 74.9 & 22.9 \\
POLPO final & 89  & 3.4 & 95.5 & 1.1 \\
\midrule
\multicolumn{5}{@{}l}{\emph{MathIF, thinking enabled}} \\
Initial     & 282 & 11.0 & 61.7 & 27.3 \\
POLPO final & 197 & 6.6 & 85.8 & 7.6 \\
\bottomrule
\end{tabular}
\end{table}

\paragraph{Dominance-pair counts during training.}
Figure~\ref{fig:if-dominance-pairs} tracks the mean number of strict dominance pairs per group, conditional on the group containing at least one such pair. From the first to the last reported window, the mean changes from 42.8 to 36.6 for IFM5 and from 42.7 to 31.4 for MulDimIF. For MathIF with thinking enabled, it changes from 14.1 in steps 1--20 to 11.9 in steps 321--325. These are conditional pair counts; they do not measure the fraction of groups that provide a learning signal. Group sizes differ across the datasets, so the maximum possible number of pairs is 120 for IFM5 and MulDimIF ($N=16$), but 28 for MathIF ($N=8$).

\paragraph{Largest incomparability component at the endpoints.}
For the initial policy and the final POLPO checkpoint, Table~\ref{tab:if-incomparability-components} summarizes the largest connected component of the incomparability graph within each group that contains a dominance pair. The graph connects responses that are incomparable under Pareto dominance (Appendix~\ref{app:general-orders}). These snapshots use responses sampled from training prompts with seed 43. Among the selected groups, the mean largest-component size changes from 8.9 to 10.7 for IFM5, from 10.9 to 11.6 for MulDimIF, and from 4.4 to 5.1 for MathIF with thinking enabled. The initial and final statistics are conditioned on their respective dominance groups; they describe the sampled group structures rather than changes within a fixed set of groups.

\begin{table}[htbp]
\centering
\small
\setlength{\tabcolsep}{4pt}
\caption{Largest incomparability-component size per group containing at least one dominance pair, for initial-policy and POLPO-final rollouts sampled from training prompts with seed 43. $N$ is the number of responses per group and $n$ is the number of groups in the conditional subset. P95 is the 95th percentile.}
\label{tab:if-incomparability-components}
\begin{tabular}{@{}lrrrrrr@{}}
\toprule
Policy & $N$ & $n$ & Mean & Median & P95 & Max \\
\midrule
\multicolumn{7}{@{}l}{\emph{IFM5}} \\
Initial     & 16 & 471 & 8.9  & 10 & 15 & 16 \\
POLPO final & 16 & 138 & 10.7 & 13 & 15 & 16 \\
\midrule
\multicolumn{7}{@{}l}{\emph{MulDimIF}} \\
Initial     & 16 & 598 & 10.9 & 12 & 15 & 16 \\
POLPO final & 16 & 89  & 11.6 & 13 & 15 & 15 \\
\midrule
\multicolumn{7}{@{}l}{\emph{MathIF, thinking enabled}} \\
Initial     & 8  & 282 & 4.4  & 4  & 7  & 8 \\
POLPO final & 8  & 197 & 5.1  & 6  & 7  & 8 \\
\bottomrule
\end{tabular}
\end{table}

\subsection{Additional MathIF Results}
\label{app:mathif-results}

With thinking disabled, POLPO achieves the highest overall Joint score among the reported methods: $21.4\%$, compared with $19.3\%$ for GRPO--Rate and $20.2\%$ for GRPO--All (Table~\ref{tab:mathif-overall}).
The following breakdowns show where these gains arise. With thinking enabled, POLPO reaches $33.8\%$ Joint, compared with $30.7\%$ for GRPO--Rate and $22.6\%$ for GRPO--All. The non-thinking breakdowns are discussed first; the thinking-mode breakdowns follow in Tables~\ref{tab:mathif-by-k-thinking} and~\ref{tab:mathif-by-source-thinking}.

\paragraph{Satisfying multiple constraints while answering correctly.}
Table~\ref{tab:mathif-by-k} shows that POLPO achieves the highest complete constraint satisfaction (HAcc) at every constraint count.
For two and three constraints, its Joint scores are $17.1\%$ and $17.9\%$, compared with $13.6\%$ and $14.3\%$ for GRPO--Rate.
With three constraints, both methods have the same mathematical accuracy at the reported precision ($27.1\%$), but POLPO has higher Joint success.
Thus, more of POLPO's mathematically correct responses also satisfy every instruction constraint.
POLPO matches GRPO--All in Joint success at one constraint and exceeds it at two and three constraints. GRPO--All has higher mathematical accuracy at two and three constraints, while GRPO--Rate leads on accuracy at one constraint.
These results show an advantage in meeting the combined requirements of correctness and instruction adherence.
The advantage over GRPO--Rate is concentrated in the multiple-constraint settings: with one constraint, POLPO's Joint score is slightly lower ($29.3\%$ versus $30.0\%$).

\paragraph{Results across benchmark sources.}
Table~\ref{tab:mathif-by-source} shows that POLPO exceeds GRPO--Rate in pooled Joint success on all five benchmark sources and achieves or ties the highest score on four of them.
On MATH-500, POLPO reaches $37.8\%$, compared with $35.6\%$ for GRPO--Rate and $33.3\%$ for GRPO--All; on Minerva, it reaches $8.9\%$, compared with $5.6\%$ and $6.7\%$, respectively.
On OlympiadBench, POLPO reaches $13.3\%$, compared with $12.2\%$ for GRPO--All and $11.1\%$ for GRPO--Rate.
On AIME, only POLPO obtains a joint success, although this amounts to just one of the 60 examples.
GSM8K is the exception to POLPO's leading pooled score: its $38.9\%$ exceeds GRPO--Rate ($37.8\%$) but remains below GRPO--All ($42.2\%$).
The improvements over GRPO--Rate therefore occur across sources, while the comparison with GRPO--All depends on the source.

\begin{table}[t]
\centering
\small
\setlength{\tabcolsep}{3pt}
\caption{MathIF results (\%) by constraint count, with 140 examples per count. Within each metric, columns are the initial model (Init.), GRPO--All (All), GRPO--Rate (Rate), and POLPO (POLPO). All models use non-thinking mode.}
\label{tab:mathif-by-k}
\begin{tabular}{c*{3}{rrrr}}
\toprule
& \multicolumn{4}{c}{HAcc} & \multicolumn{4}{c}{Acc} & \multicolumn{4}{c}{Joint} \\
\cmidrule(lr){2-5}\cmidrule(lr){6-9}\cmidrule(lr){10-13}
$K$ & Init. & All & Rate & POLPO & Init. & All & Rate & POLPO & Init. & All & Rate & POLPO \\
\midrule
1 & 56.4 & 78.6 & 85.7 & 89.3 & 15.0 & 32.9 & 33.6 & 30.7 & 10.7 & 29.3 & 30.0 & 29.3 \\
2 & 22.9 & 50.0 & 69.3 & 78.6 & 10.0 & 25.7 & 20.7 & 22.9 & 2.9 & 15.7 & 13.6 & 17.1 \\
3 & 14.3 & 38.6 & 72.9 & 76.4 & 16.4 & 35.7 & 27.1 & 27.1 & 3.6 & 15.7 & 14.3 & 17.9 \\
\bottomrule
\end{tabular}
\end{table}

\begin{table}[t]
\centering
\small
\caption{MathIF joint success (\%) by source and constraint count. Joint success requires mathematical correctness and satisfaction of every instruction constraint. All models use non-thinking mode. Rows marked All pool the three constraint counts for that source.}
\label{tab:mathif-by-source}
\begin{tabular}{lrrrrrr}
\toprule
Source & $K$ & $n$ & Initial & GRPO--All & GRPO--Rate & POLPO \\
\midrule
GSM8K & 1 & 30 & 26.7 & 56.7 & 60.0 & 63.3 \\
GSM8K & 2 & 30 & 10.0 & 33.3 & 23.3 & 23.3 \\
GSM8K & 3 & 30 & 6.7 & 36.7 & 30.0 & 30.0 \\
GSM8K & All & 90 & 14.4 & 42.2 & 37.8 & 38.9 \\
\midrule
MATH-500 & 1 & 30 & 16.7 & 50.0 & 60.0 & 53.3 \\
MATH-500 & 2 & 30 & 3.3 & 30.0 & 26.7 & 33.3 \\
MATH-500 & 3 & 30 & 3.3 & 20.0 & 20.0 & 26.7 \\
MATH-500 & All & 90 & 7.8 & 33.3 & 35.6 & 37.8 \\
\midrule
Minerva & 1 & 30 & 6.7 & 10.0 & 6.7 & 6.7 \\
Minerva & 2 & 30 & 0.0 & 6.7 & 6.7 & 13.3 \\
Minerva & 3 & 30 & 3.3 & 3.3 & 3.3 & 6.7 \\
Minerva & All & 90 & 3.3 & 6.7 & 5.6 & 8.9 \\
\midrule
OlympiadBench & 1 & 30 & 0.0 & 20.0 & 13.3 & 13.3 \\
OlympiadBench & 2 & 30 & 0.0 & 3.3 & 6.7 & 10.0 \\
OlympiadBench & 3 & 30 & 3.3 & 13.3 & 13.3 & 16.7 \\
OlympiadBench & All & 90 & 1.1 & 12.2 & 11.1 & 13.3 \\
\midrule
AIME & 1 & 20 & 0.0 & 0.0 & 0.0 & 0.0 \\
AIME & 2 & 20 & 0.0 & 0.0 & 0.0 & 0.0 \\
AIME & 3 & 20 & 0.0 & 0.0 & 0.0 & 5.0 \\
AIME & All & 60 & 0.0 & 0.0 & 0.0 & 1.7 \\
\bottomrule
\end{tabular}
\end{table}

\paragraph{Thinking-mode breakdowns.}
With thinking enabled, POLPO has the highest HAcc at each constraint count among the methods shown in Table~\ref{tab:mathif-by-k-thinking}. Its Joint scores are $40.7\%$, $30.7\%$, and $30.0\%$ for one, two, and three constraints, respectively, compared with $40.0\%$, $28.6\%$, and $27.1\%$ for DPO--Pareto and $37.1\%$, $27.9\%$, and $27.1\%$ for GRPO--Rate. Across sources, POLPO leads the methods shown on MATH-500 and OlympiadBench; DPO--Pareto leads on GSM8K and AIME, and GRPO--Rate leads on Minerva. The source subsets are small, especially AIME ($n=60$).

\begin{table}[t]
\centering
\small
\setlength{\tabcolsep}{3pt}
\caption{MathIF results (\%) by constraint count $K$ with thinking enabled and a 16,384-token evaluation response limit, with 140 examples per count. HAcc measures complete constraint satisfaction, Acc measures mathematical correctness, and Joint requires both. Bold indicates the best result among the methods shown in each column.}
\label{tab:mathif-by-k-thinking}
\resizebox{\linewidth}{!}{%
\begin{tabular}{lrrrrrrrrr}
\toprule
& \multicolumn{3}{c}{HAcc} & \multicolumn{3}{c}{Acc} & \multicolumn{3}{c}{Joint} \\
\cmidrule(lr){2-4}\cmidrule(lr){5-7}\cmidrule(lr){8-10}
Method & $K=1$ & $K=2$ & $K=3$ & $K=1$ & $K=2$ & $K=3$ & $K=1$ & $K=2$ & $K=3$ \\
\midrule
Initial model & 40.0 & 22.9 & 12.9 & 31.4 & 22.1 & 21.4 & 12.9 & 3.6 & 2.1 \\
GRPO--All & 50.7 & 40.0 & 35.0 & 31.4 & 25.7 & 24.3 & 29.3 & 20.0 & 18.6 \\
GRPO--Rate & 81.4 & 70.0 & 60.0 & 42.9 & 36.4 & 37.1 & 37.1 & 27.9 & 27.1 \\
GDPO & 69.3 & 59.3 & 47.1 & 33.6 & 25.7 & 32.1 & 29.3 & 20.0 & 20.7 \\
GraphDPO & 68.6 & 55.0 & 45.7 & 33.6 & 25.7 & 30.7 & 31.4 & 21.4 & 21.4 \\
DPO--Pareto & 87.1 & 74.3 & 56.4 & 45.0 & \textbf{37.1} & \textbf{38.6} & 40.0 & 28.6 & 27.1 \\
\rowcolor{polpohighlight}
POLPO & \textbf{87.9} & \textbf{79.3} & \textbf{67.1} & \textbf{45.7} & \textbf{37.1} & 37.9 & \textbf{40.7} & \textbf{30.7} & \textbf{30.0} \\
\bottomrule
\end{tabular}}
\end{table}

\begin{table}[t]
\centering
\small
\setlength{\tabcolsep}{3pt}
\caption{MathIF joint success (\%) by source and constraint count with thinking enabled and a 16,384-token evaluation response limit. Joint success requires mathematical correctness and satisfaction of every instruction constraint. Rows marked All pool the three constraint counts for that source.}
\label{tab:mathif-by-source-thinking}
\resizebox{\linewidth}{!}{%
\begin{tabular}{lrrrrrrrrr}
\toprule
Source & $K$ & $n$ & Initial & GRPO--All & GRPO--Rate & GDPO & GraphDPO & DPO--Pareto & POLPO \\
\midrule
GSM8K & 1 & 30 & 30.0 & 46.7 & 73.3 & 53.3 & 50.0 & 70.0 & 60.0 \\
GSM8K & 2 & 30 & 10.0 & 33.3 & 50.0 & 33.3 & 40.0 & 53.3 & 46.7 \\
GSM8K & 3 & 30 & 3.3 & 20.0 & 36.7 & 33.3 & 26.7 & 43.3 & 50.0 \\
GSM8K & All & 90 & 14.4 & 33.3 & 53.3 & 40.0 & 38.9 & 55.6 & 52.2 \\
\midrule
MATH-500 & 1 & 30 & 16.7 & 60.0 & 63.3 & 56.7 & 63.3 & 63.3 & 70.0 \\
MATH-500 & 2 & 30 & 3.3 & 30.0 & 40.0 & 33.3 & 36.7 & 46.7 & 56.7 \\
MATH-500 & 3 & 30 & 0.0 & 40.0 & 43.3 & 33.3 & 33.3 & 40.0 & 56.7 \\
MATH-500 & All & 90 & 6.7 & 43.3 & 48.9 & 41.1 & 44.4 & 50.0 & 61.1 \\
\midrule
Minerva & 1 & 30 & 3.3 & 6.7 & 13.3 & 3.3 & 13.3 & 16.7 & 6.7 \\
Minerva & 2 & 30 & 3.3 & 10.0 & 16.7 & 10.0 & 6.7 & 10.0 & 10.0 \\
Minerva & 3 & 30 & 3.3 & 3.3 & 13.3 & 3.3 & 6.7 & 3.3 & 3.3 \\
Minerva & All & 90 & 3.3 & 6.7 & 14.4 & 5.6 & 8.9 & 10.0 & 6.7 \\
\midrule
OlympiadBench & 1 & 30 & 6.7 & 20.0 & 20.0 & 20.0 & 16.7 & 30.0 & 40.0 \\
OlympiadBench & 2 & 30 & 0.0 & 10.0 & 16.7 & 6.7 & 10.0 & 13.3 & 23.3 \\
OlympiadBench & 3 & 30 & 3.3 & 16.7 & 30.0 & 26.7 & 30.0 & 33.3 & 30.0 \\
OlympiadBench & All & 90 & 3.3 & 15.6 & 22.2 & 17.8 & 18.9 & 25.6 & 31.1 \\
\midrule
AIME & 1 & 20 & 5.0 & 5.0 & 5.0 & 5.0 & 5.0 & 10.0 & 20.0 \\
AIME & 2 & 20 & 0.0 & 15.0 & 10.0 & 15.0 & 10.0 & 15.0 & 10.0 \\
AIME & 3 & 20 & 0.0 & 10.0 & 5.0 & 0.0 & 5.0 & 10.0 & 0.0 \\
AIME & All & 60 & 1.7 & 10.0 & 6.7 & 6.7 & 6.7 & 11.7 & 10.0 \\
\bottomrule
\end{tabular}}
\end{table}

%

\FloatBarrier
\section{Extended Related Work}
\label{app:extended-related}

This appendix complements Section~\ref{sec:related} with related optimization and evaluation methods beyond the direct comparison between scalar aggregation and partial-order likelihoods in multi-verifier RLVR.

\subsection{Additional Approaches to Multiple Verification Signals}
\label{app:related-multi-reward}

IFEval provides automatically checkable instruction constraints that underpin later work on verifiable instruction following \citep{zhou2023ifeval}. VerIF combines code-based checks for hard constraints with language-model judgments for soft constraints, then aggregates their outcomes for GRPO training \citep{peng2025verif}; its feedback therefore includes learned judgments as well as executable checks. MDP-GRPO uses the mean constraint-satisfaction reward and modifies sampling and advantage estimation to address difficult reward distributions in multi-constraint instruction following \citep{salmanizarchi2026mdpgrpo}.

Beyond the aggregation methods discussed in the main text, GD$^2$PO addresses conflicts between component advantages through rollout masking and prompt reweighting \citep{liu2026gd2po}. Blockwise advantage estimation applies objective-specific advantages to the corresponding output segments \citep{pavlenko2026blockwise}. These approaches show that using multiple rewards need not mean assigning one undifferentiated signal to every token. POLPO addresses a different design choice: it uses comparisons between complete responses, without assuming a correspondence between verification criteria and output segments.

\subsection{Policy Optimization and Listwise RLVR}
\label{app:related-policy}

Reinforcement learning from human feedback trains policies using learned preference rewards \citep{ouyang2022instructgpt}, while RLVR obtains feedback from automatic verification \citep{lambert2024tulu3,deepseek2025r1}. GRPO constructs advantages from rewards within a response group \citep{shao2024deepseekmath} and uses the clipped policy surrogate of PPO \citep{schulman2017ppo}. See Appendix~\ref{app:grpo-gradient} for the definitions needed for the gradient comparison in this paper.

Listwise Policy Optimization interprets group-based RLVR as projection onto target distributions over sampled responses \citep{qu2026lpo}. Its targets are constructed from scalar rewards, and its update coefficients can depend on current policy scores and sum to zero within the group. These properties are therefore not unique to POLPO. The distinction is the source of the coefficients: POLPO differentiates the probability of a verifier-induced partial-order event, marginalizing over compatible complete rankings rather than projecting onto a scalar-reward target distribution.

\subsection{Listwise and Set-Level Preference Optimization}
\label{app:related-ranking}

DPO uses policy-to-reference log-probability ratios in a Bradley--Terry comparison model and extends this construction to complete rankings with the Plackett--Luce model \citep{rafailov2023dpo}. The complete-ranking negative log-likelihood is the ListMLE loss \citep{xia2008listmle}; ListNet provides another listwise learning-to-rank formulation \citep{cao2007listnet}. PRO \citep{song2024pro} and LiPO \citep{liu2025lipo} bring learning from ranked or scored response lists to language-model alignment. POLPO instead specifies a partial-order event and sums the probabilities of the complete rankings that satisfy it.

Related methods also learn from sets rather than individual pairs. In recommendation, S-DPO contrasts one positive item with several negatives, corresponding to the PL probability that the positive is selected first \citep{chen2024sdpo}. Mult-DPO handles multiple positives, negatives, and preference levels with a multinomial surrogate that upper-bounds the corresponding marginalized PL loss \citep{zhu2026multdpo}. Multi-Preference Optimization generalizes Bradley--Terry comparisons to chosen and rejected sets \citep{gupta2024mpo}. These set-based objectives address particular comparison structures.
The PL likelihood for a general partial order, formed by summing
over compatible complete rankings, is established in the ranking
literature \citep{liu2019partial}; ordered-partition likelihoods
have also been studied explicitly \citep{ma2021partitioned}.
POLPO adopts this probabilistic construction for verifier-induced
comparisons between model responses and optimizes it through
DPO-style policy scores in an online sampling-and-update procedure.

\subsection{Verifier-Guided Preferences and Prompt Optimization}
\label{app:related-verifier-preferences}

Verifier outcomes can also generate pairwise preference data for reasoning. Iterative Reasoning Preference Optimization pairs correct and incorrect responses and adds a negative log-likelihood term on the correct response \citep{pang2024irpo}. DPO-VP constructs correct--incorrect pairs from binary verifiable rewards and trains iteratively \citep{tu2025iterativedpo}. These methods connect verification to preference optimization, while POLPO uses the comparison structure of a response group and models its specified comparisons as one ranking event.

ParetoPrompt uses dominance relations between prompts for preference-based policy optimization \citep{zhao2025paretoprompt}. This is related through its use of multiple criteria, but its optimization objects are prompts. POLPO constructs comparisons between responses to the same prompt and updates the response policy through their joint partial-order likelihood.

\end{document}

%% file: math_commands.tex
\usepackage{amsmath,amsfonts,bm}

\def\eqref#1{equation~\ref{#1}}

\def\1{\bm{1}}

\DeclareMathAlphabet{\mathsfit}{\encodingdefault}{\sfdefault}{m}{sl}
\SetMathAlphabet{\mathsfit}{bold}{\encodingdefault}{\sfdefault}{bx}{n}

\newcommand{\E}{\mathbb{E}}

